\documentclass[3p]{elsarticle}

\usepackage{amsmath,amssymb,amsthm}
\usepackage{booktabs}

\newtheorem{theorem}{Theorem}
\newtheorem{proposition}[theorem]{Proposition}%
\newtheorem{lemma}{Lemma}%
\newtheorem{corollary}{Corollary}%
\newtheorem{example}{Example}%

\usepackage{multirow}
\usepackage{url}

\usepackage{xcolor}

\usepackage[all]{xy}
\usepackage{tikz}
\usetikzlibrary{arrows.meta}
\usetikzlibrary{positioning,shapes,arrows}

\usepackage{pifont}
\newcommand{\cmark}{\ding{51}}%
\newcommand{\xmark}{\ding{55}}%

\begin{document}
\begin{frontmatter}

\title{Embeddings as Epistemic States: Limitations on the Use of Pooling Operators for Accumulating Knowledge}

\author{Steven Schockaert}
\ead{schockaerts1@cardiff.ac.uk}

\address{Cardiff University, School of Computer Science \& Informatics,\\Abacws building, Senghennydd Road, CF24 4AG, Cardiff, UK}

\begin{abstract}
Various neural network architectures rely on pooling operators to aggregate information coming from different sources. It is often implicitly assumed in such contexts that vectors encode epistemic states, i.e.\ that vectors capture the evidence that has been obtained about some properties of interest, and that pooling these vectors yields a vector that combines this evidence. We study, for a number of standard pooling operators, under what conditions they are compatible with this idea, which we call the epistemic pooling principle. While we find that all the considered pooling operators can satisfy the epistemic pooling principle, this only holds when embeddings are sufficiently high-dimensional and, for most pooling operators, when the embeddings satisfy particular constraints (e.g.\ having non-negative coordinates). 
We furthermore show that these constraints have important implications on how the embeddings can be used in practice. In particular,
we find that when the epistemic pooling principle is satisfied, in most cases it is impossible to verify the satisfaction of propositional formulas using linear scoring functions, with two exceptions: (i) max-pooling with embeddings that are upper-bounded and (ii) Hadamard pooling with non-negative embeddings. This finding helps to clarify, among others, why Graph Neural Networks sometimes under-perform in reasoning tasks.
Finally, we also study an extension of the epistemic pooling principle to weighted epistemic states, which are important in the context of non-monotonic reasoning, where max-pooling emerges as the most suitable operator.
\end{abstract}

\end{frontmatter}

\section{Introduction}
One of the key challenges in many sub-areas of Machine Learning is to learn suitable vector space embeddings of the objects of interest (e.g.\ graphs, images or sentences). A question which is usually left implicit is what the embedding of an object represents. We can take at least two different views on this. First, we may consider that embeddings essentially serve as a compact representation of a \emph{distance metric}, which intuitively captures some form of similarity. What matters, then, is that objects which are similar, in some sense, are represented by similar vectors, while objects which are dissimilar are not. This intuition provides the foundation, for instance, for the use of contrastive pre-training strategies \cite{DBLP:conf/cvpr/ChopraHL05,DBLP:journals/corr/abs-1807-03748}. Second, we may consider that embeddings are essentially compact encodings of \emph{epistemic states}. In other words, the embedding of an object encodes what we know about that object. 
Embeddings then essentially play a similar role as formulas in propositional logic.
This view implicitly underpins most strategies that combine neural network learning with aspects of symbolic reasoning, e.g.\ when using a semantic loss function to encourage neural network predictions to satisfy certain constraints \cite{DBLP:conf/icml/XuZFLB18} or when using neural network predictions as input to a probabilistic logic program \cite{DBLP:conf/nips/ManhaeveDKDR18}. In this paper, we focus on this second view. 

In practice, the embedding of an object is often obtained by combining the embeddings of related objects using some kind of pooling operator. For instance, in the context of Computer Vision, convolutional feature extractors such as ResNet \cite{residual} provide an embedding for each sub-region of the image. An embedding of the overall image is then typically obtained by averaging these sub-region embeddings. Along similar lines, a common setting in Natural Language Processing consists in using a transformer based language model such as BERT \cite{BERT} to obtain paragraph embeddings, and to average these embeddings to obtain an embedding for a full document. In multi-modal settings, it is common to obtain embeddings for the individual modalities first, and to subsequently aggregate these embeddings \cite{DBLP:conf/aaai/KielaGJM18}. Graph Neural Networks \cite{scarselli2008graph,schlichtkrull2018modeling} also crucially rely on pooling operators, learning node representations by aggregating embeddings derived from neighbouring nodes.  Essentially, in all these cases we have an embedding $\mathbf{e}$ which is obtained by aggregating embeddings $\mathbf{e_1},...,\mathbf{e_k}$ using some pooling operator $\diamond$:
\begin{align}\label{eqPoolingPrincipleA}
\mathbf{e} = \diamond(\mathbf{e_1},...,\mathbf{e_k})
\end{align}
Under the epistemic\footnote{The word \emph{epistemic}, in this paper, merely refers to the idea that vectors capture what we know about some properties of interest. In particular, note that our setting is not directly related to epistemic logics.} view, this pooling operator is implicitly assumed to aggregate the knowledge that is captured by the embeddings $\mathbf{e_1},...,\mathbf{e_k}$. For instance, the embeddings $\mathbf{e_1},...,\mathbf{e_k}$ may encode which objects are present in different parts of the image. After pooling these embeddings, we should end up with an embedding $\mathbf{e}$ that captures which objects are present throughout the entire image. Let us write $\Gamma(\mathbf{e_i})$ for the knowledge that is captured by the embedding $\mathbf{e_i}$. More precisely, we will think of $\Gamma(\mathbf{e_i})$ as a set of properties that are known to be satisfied. If we view pooling as the process of accumulating knowledge from different sources (e.g.\ from different regions of the image, or different neighbours in a graph neural network), then we would expect the following to be true: $\Gamma(\mathbf{e}) = \Gamma(\mathbf{e_1})\cup ... \cup \Gamma(\mathbf{e_n})$. We will refer to this principle as the \emph{epistemic pooling principle}. As an important special case, we will consider the case where the properties of interest correspond to possible worlds (i.e.\ propositional interpretations). The set $\Gamma(\mathbf{e_i})$ then contains the set of possible worlds that can be excluded based on the knowledge encoded in the embedding $\mathbf{e_i}$. In other words, $\Gamma(\mathbf{e_i})$ can then be characterised as a propositional formula. The epistemic pooling principle then states that the formula corresponding to $\diamond(\mathbf{e_1},...,\mathbf{e_k})$ should be equivalent to the conjunction of the formulas corresponding to $\mathbf{e_1},...,\mathbf{e_k}$.

The main aim of this paper is to study under which conditions the epistemic pooling principle can be satisfied. Analysing a number of standard pooling operators, we find that the epistemic pooling principle can be satisfied for all of them, but with several important caveats:
\begin{itemize}
\item We need at least as many dimensions as there are properties. In settings where we want to model propositional formulas, the properties of interest correspond to possible worlds. Without further restrictions, this means that we need embeddings with as many dimensions as there are possible worlds.
\item For most of the pooling operators, we find that embeddings need to be constrained in a particular way, e.g.\ by only allowing vectors with non-negative coordinates.
\item We also identify important restrictions on how embeddings can be linked to the formulas they capture. In particular, when summation or averaging is used for pooling, we find that the satisfaction of propositional formulas can, in general, not be predicted by a linear classifier when the epistemic pooling principle is satisfied.
\end{itemize}
The fundamental question which we want to answer is whether a vector-based representation can act as a formal knowledge representation framework, or whether reasoning with neural networks is inevitably approximate in nature. 
While we focus on a theoretical analysis of pooling operators in this paper, our results provide a number of important insights for the design of neural network models for tasks that require reasoning. For instance, two operators emerge from our analysis as being particularly suitable for applications where we need to reason about propositional formulas: max-pooling, with the constraint that the coordinates of all embeddings are upper-bounded by some constant $z$, and the Hadamard operator (i.e.\ the component-wise product), with the constraint that all coordinates are non-negative. 

Our results also help to explain the outperformance of neuro-symbolic methods \cite{DBLP:series/faia/BesoldGBBDHKLLPPPZ21} over Graph Neural Networks (GNN) in certain applications. GNNs are, in theory, capable of reasoning in the two-variable fragment of first-order logic with counting quantifiers (FOC\textsubscript{2}) \cite{DBLP:conf/iclr/BarceloKM0RS20}. Nonetheless, in practice, GNNs often perform relatively poorly in tasks that require reasoning, even when staying within FOC\textsubscript{2}. Crucially, GNNs typically use averaging (or summation) for pooling the messages coming from adjacent nodes. As we will see in Section \ref{secReasoningWithAverages}, in such cases, GNNs can only capture logical reasoning 
if the node embeddings are essentially binary. The discrete nature of the required embeddings makes them difficult to learn, which helps to explain why GNNs are often outperformed by models that are more specifically tailored towards reasoning \cite{DBLP:conf/icml/Minervini0SGR20,DBLP:conf/nips/BergenOB21}. 

Furthermore, our lower bounds on the required dimensionality of epistemic embeddings suggest that, beyond toy problems, using neural networks for reasoning inevitably requires some kind of modularity. For instance, an important challenge in Natural Language Processing is to design models that can reason about information that comes from different sources (e.g.\ different news sources). Combining different text fragments by pooling their embeddings would require a prohibitively high dimensionality, if we want these embeddings to capture epistemic states in a faithful way. Instead, it is more common to use vectors to encode what we know about a given entity, or about the relationship between two entities. Reasoning about the overall story then requires us to combine these different entity and relation vectors, by using the structure of the problem domain in some way, e.g.\ by using GNNs or neuro-symbolic approaches. By exploiting this structure, embeddings can be used to capture more focused knowledge, which means that the prohibitively high dimensionality that is otherwise needed can be avoided.

The remainder of this paper is structured as follows. In the next section, we first discuss a number of simple examples to illustrate the considered setting. Subsequently, in Section \ref{secProblemSetting}, we formalise the problem setting and introduce the notations that will be used throughout the paper. In Section \ref{secPoolingProperties}, we then analyse under which conditions the epistemic pooling principle can be satisfied. One of the main findings from this section is that the requirement to satisfy the epistemic pooling principle fundamentally constrains which embeddings can be allowed, and how these embeddings encode knowledge. In Section \ref{secPropositionalReasoning}, we then investigate how this impacts our ability to use vector embeddings for propositional reasoning. In particular, we focus on the problem of verifying whether a given propositional formula is satisfied in the epistemic state encoded by a given vector. Subsequently, in Section \ref{secWeightedEpistemicStates}, we look at a generalisation of the epistemic pooling principle for dealing with weighted epistemic states. In this case, vectors encode the strength with which we believe a given property to be satisfied. Section \ref{secRelatedWork} presents a discussion of our results in the context of related work, after which we summarise our conclusions.

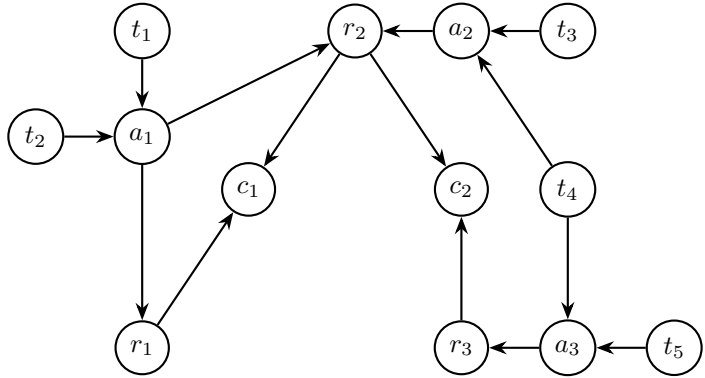
\begin{figure}
\centering
\begin{tikzpicture}[scale=0.7]
\begin{scope}[every node/.style={circle,thick,draw}]
    \node (C1) at (0,0) {$c_1$};
    \node (C2) at (4,0) {$c_2$};
    \node (R1) at (-2,-3) {$r_1$};
    \node (R2) at (2,3) {$r_2$};
    \node (R3) at (4,-3) {$r_3$};
    \node (A1) at (-2,1) {$a_1$};
    \node (A2) at (4,3) {$a_2$};
    \node (A3) at (6,-3) {$a_3$};
    \node (T1) at (-2,3) {$t_1$};
    \node (T2) at (-4,1) {$t_2$};
    \node (T3) at (6,3) {$t_3$};
    \node (T4) at (6,0) {$t_4$};
    \node (T5) at (8,-3) {$t_5$};
\end{scope}

\begin{scope}[>={Stealth[black]},              
            every edge/.style={draw=black,thick}]   
            \path [->] (R1) edge  (C1);   
            \path [->] (R2) edge  (C1);             
            \path [->] (R2) edge  (C2);
            \path [->] (R3) edge  (C2);
            \path [->] (A1) edge  (R1);
            \path [->] (A1) edge  (R2);
            \path [->] (A2) edge  (R2);
            \path [->] (A3) edge  (R3);
            \path [->] (T1) edge  (A1);
            \path [->] (T2) edge  (A1);
            \path [->] (T3) edge  (A2);
            \path [->] (T4) edge  (A2);
            \path [->] (T4) edge  (A3);
            \path [->] (T5) edge  (A3);
\end{scope}
\end{tikzpicture}
\caption{A graph containing four types of nodes. The nodes $c_1,c_2$ represent committees, $r_1,r_2,r_3$ represent researchers, $a_1,a_2,a_3$ represent articles, and $t_1,t_2,t_3,t_4,t_5$ represent research topics. Topic nodes are connected to the articles that discuss them. Article nodes are connected to their authors. Researchers are connected to the committees they belong to.\label{figIllustrativeExample1}}
\end{figure}

\section{Illustrative examples}
In this section, we consider a number of toy examples to illustrate our problem setting. These examples involve message-passing GNNs, where pooling arises because we need to aggregate the messages coming from adjacent nodes. In Section \ref{secExampleBasic}, we discuss a simple example to illustrate the basic principles. This example involves aggregating sets of properties, without any further reasoning. We then build on this example in Section \ref{secExPropositional} to illustrate how propositional reasoning can be carried out in this framework. Section \ref{secExBackground} illustrates a setting where we need to reason in the presence of background knowledge. Finally, Section \ref{secExNMR} shows how this can be extended to situations involving non-monotonic reasoning with defeasible knowledge.

\subsection{Basic setting} \label{secExampleBasic}
Figure \ref{figIllustrativeExample1} displays a graph with four types of nodes, referring to research topics ($t_1,t_2,t_3,t_4,t_5$), scientific articles ($a_1,a_2,a_3$), researchers ($r_1,r_2,r_3$) and committees ($c_1,c_2$), respectively. In this graph, research topics are connected to the articles that discuss them, articles are connected to their authors and researchers are connected to the committees they belong to.
We say that a researcher is an expert on a topic if they have published at least one article which discusses this topic. We say that a committee is complete if it contains an expert on each of the five topics. For instance, in the case of Figure \ref{figIllustrativeExample1}, we can see that researcher $r_1$ is an expert on topics $t_1,t_2$; researcher $r_2$ is an expert on topics $t_1,t_2,t_3,t_4$; and researcher $r_3$ is an expert on topics $t_4,t_5$. It follows that committee $c_2$ is complete, as each of the five topics are covered by its two members, but committee $c_1$ is not. We are now interested in designing a message-passing GNN which can predict whether a given committee is complete or not. A message-passing GNN learns embeddings of the nodes of a given graph, by iteratively updating each node's embedding based on the embeddings of its neighbouring nodes. To specify a message-passing GNN, we thus need to specify (i) the initial embedding of each node and (ii) the update mechanism which is used.

A straightforward solution is as follows. Let us write $\mathbf{x^{(l)}}\in \mathbb{R}^5$ for the representation of a node $x$ in layer $l$ of the GNN. In particular, $\mathbf{x^{(0)}}$ represents the input embedding of node $x$, which we define as follows:
\begin{align*}
\mathbf{t_1^{(0)}}&=(1,0,0,0,0) &
\mathbf{t_2^{(0)}}&=(0,1,0,0,0) &
\mathbf{t_3^{(0)}}&=(0,0,1,0,0) &
\mathbf{t_4^{(0)}}&=(0,0,0,1,0) &
\mathbf{t_5^{(0)}}&=(0,0,0,0,1) &
\end{align*}
In other words, we use a one-hot encoding for representing the different topics. The input embeddings for all the other nodes are set to $\mathbf{x^{(0)}}=(0,0,0,0,0)$. In the subsequent layers, the node representations are updated as follows:
\begin{align}\label{eqExGNNUpdate}
\mathbf{x^{(l+1)}} = \max(\{\mathbf{x^{(l)}}\}\cup\{\mathbf{y^{(l)}} \mid (y,x)\in\mathcal{E} \})
\end{align}
where $\mathcal{E}$ represents the set of edges, i.e.\ $(y,x)\in\mathcal{E}$ if there is an edge from node $y$ to node $x$, and the maximum is applied component-wise. It is easy to verify that an article $a_i$ covers topic $t_j$ if the $j\textsuperscript{th}$ coordinate of $\mathbf{a_i^{(l)}}$ is $1$, for $l\geq 1$. Similarly, a researcher $r_i$ is an expert on topic $j$ if the $j\textsuperscript{th}$ coordinate of $\mathbf{r_i^{(l)}}$ is $1$, for $l\geq 2$; and a committee $c_i$ contains expertise on topic $j$ if the $j\textsuperscript{th}$ coordinate of $\mathbf{c_i^{(l)}}$ is $1$, for $l\geq 3$. We thus have that the committee $c_i$ is complete iff $\mathbf{c_i^{(3)}}=(1,1,1,1,1)$. Note that the latter condition can be checked using a linear scoring function, since it is equivalent with
$\mathbf{c_i}^{(3)} \cdot \mathbf{1} \geq 5$, where $\mathbf{1}=(1,1,1,1,1)$.

In the proposed construction, a number of particular design decisions were made. For instance, we used 5-dimensional embeddings for representing nodes and we used the maximum for aggregating the evidence from neighbouring nodes. Our analysis in this paper is aimed at studying the importance of such decisions. For instance, we may wonder whether it is possible to devise an encoding that relies on four-dimensional embeddings. If we are allowed to replace the maximum with an arbitrarily complex function (and we allow non-linear scoring functions), then the answer is clearly positive\footnote{In fact, we can even use one-dimensional embeddings. Let $n$ be a number which is larger than the number of topics $k$. Then we can initialise topic $i$ as $n^{i-1}$. To aggregate two node representations $x$ and $y$, we first decompose them as $x= x_0 + x_1n + x_2n^2 + ... + x_{k-1}n^{k-1}$ and $y= y_0 + y_1n + y_2n^2 + ... + y_{k-1}n^{k-1}$, with $x_0,...,x_{k-1},y_0,...y_{k-1} \in \{0,1\}$. Then the aggregated embedding can be defined as $\max(x_0,y_0) + \max(x_1,y_1)n + ... + \max(x_{k-1},y_{k-1})n^{k-1}$.}. However, as we will see in Section \ref{secPoolingProperties}, when we restrict ourselves to standard pooling operators, such as max-pooling, summation or averaging, the answer is negative. Another question is whether a variant of the construction above can be found which relies on summation or averaging, rather than max-pooling. Here, the answer depends on what assumptions we make on the final node classifier, i.e.\ the function that maps the embedding $\mathbf{c^{(l)}}$ of a committee onto a decision. The most common approach is to rely on a linear classifier to make such predictions. In that case, we can show that no encoding of the problem can be found that relies on summation or averaging, as we will see in Section \ref{secPropositionalReasoning}.

\subsection{Propositional reasoning}\label{secExPropositional}
The aforementioned example simply required us to aggregate sets of features, which corresponded to research topics in that example. Throughout this paper, we will refer to such features as \emph{properties}. At first glance, it may seem like this setting only involves rather basic forms of reasoning. However, by identifying the considered properties with possible worlds, we can in fact design GNNs which perform propositional reasoning. We illustrate this with an example. Let us consider the same setting as before, but now we only focus on research topics, articles and researchers. We say that a researcher is a generalist if they have worked on at least two sufficiently distinct topics. Let us write $t_i$ to denote that some researcher has worked on topic $t_i$. Then we define:
$$
\textit{generalist} \equiv (t_1 \wedge t_4) \vee (t_1 \wedge t_5) \vee (t_2 \wedge t_5) 
$$
where the idea is that the other topic combinations are too similar to each other (e.g.\ working on $t_1$ and $t_2$ would not make someone a generalist because these are related research topics). We want to predict whether a given researcher is a generalist by applying a linear classifier to the corresponding node embedding. As before, we assume that available knowledge about researchers is encoded as a graph with topic, article and researcher nodes (although we no longer consider committees). Let $\omega_1,...,\omega_{32}$ be an enumeration of all the possible worlds of the propositional logic over the set of atoms $\{t_1,...,t_5\}$. The input embedding $\mathbf{t_i^{(0)}}$ of the node $t_i$ is now defined as a 32-dimensional vector, where the $j\textsuperscript{th}$ component is 0 if $\omega_j\models \neg t_i$ and -1 otherwise. Note how each component now corresponds to a possible world. The $j\textsuperscript{th}$ coordinate in the embedding of node $n$ is 0 if that node captures the knowledge that the world $\omega_j$ can be excluded. The input embeddings of the article and research nodes are initialised as $(-1,...,-1)$. The embeddings of the nodes in the subsequent layers are again computed using \eqref{eqExGNNUpdate}. For $l\geq 1$ we then have that $\mathbf{a_i^{(l)}}=(x_1,...,x_{32})$ where $x_j=0$ iff article $a_i$ has some topic $t_k$ such that $\omega_j\models \neg t_k$, and $x_j=-1$ otherwise. For $l\geq 2$, we have that $\mathbf{r_i^{(l)}}=(x_1,...,x_{32})$ where $x_j=0$ iff researcher $r_i$ has written an article that has some topic $t_k$ such that $\omega_j\models \neg t_k$. In other words, we have that $x_j=0$ if we can exclude $\omega_j$ as a possible representation of the expertise of $r_i$ based on the knowledge encoded in the graph. To test whether we can entail that $r_i$ is a generalist, it suffices to check whether every countermodel of $(t_1 \wedge t_4) \vee (t_1 \wedge t_5) \vee (t_2 \wedge t_5)$ can be excluded. Let $\mathbf{c}\in \mathbb{R}^{32}$ be the vector whose $j\textsuperscript{th}$ component is 0 if $\omega_i$ is a model of $(t_1 \wedge t_4) \vee (t_1 \wedge t_5) \vee (t_2 \wedge t_5)$ and 1 otherwise. Then we have that researcher $r_i$ is a generalist iff 
$$ 
\mathbf{r_i^{(2)}} \cdot \mathbf{c} \geq 0
$$
By associating properties with possible worlds, we can thus study settings in which GNNs need to aggregate knowledge encoded using propositional formulas. While the kind of knowledge that we had to consider in this example was rather simple, in general the formulas that need to be aggregated can be arbitrarily complex propositional formulas (including e.g.\ formulas with negation). The topic of propositional reasoning by pooling embeddings, as illustrated in the aforementioned example, will be studied in Section \ref{secPropositionalReasoning}. Among others, we will see that the choice of the pooling operator plays an important role, where faithful reasoning is, in general, not possible with averaging and summation, when linear scoring functions are used.

\subsection{Background knowledge}\label{secExBackground} Background knowledge can straightforwardly be taken into account by restricting the set of possible worlds to the models of a given knowledge base. Embeddings then capture which models of the knowledge base are still possible. To illustrate this, let us consider a similar setting as before, but with four topics: \emph{Artificial Intelligence}, \emph{Machine Learning}, \emph{Knowledge Representation} and \emph{Databases}. Let us denote these topics as $t_{\text{AI}}$, $t_{\text{ML}}$, $t_{\text{KR}}$ and $t_{\text{DB}}$ and suppose we have a knowledge base $K$ containing the following two rules:
\begin{align}
t_{\text{ML}} &\rightarrow t_{\text{AI}} \label{eqExBackgroundKnowledge1}\\
t_{\text{KR}} &\rightarrow t_{\text{AI}} \label{eqExBackgroundKnowledge2}
\end{align}
In other words, if an article is about Machine Learning, then it is also about Artificial Intelligence, and the same holds for Knowledge Representation.
This knowledge base has the following models (where we denote an interpretation by the set of atoms it makes true):
\begin{align*}
\omega_1&=\{\} &
\omega_2&=\{t_{\text{AI}}\} &
\omega_3&=\{t_{\text{AI}},t_{\text{ML}}\} &
\omega_4&=\{t_{\text{AI}},t_{\text{KR}}\} &
\omega_5&=\{t_{\text{AI}},t_{\text{ML}},t_{\text{KR}}\} \\
\omega_6&=\{t_{\text{DB}}\} &
\omega_7&=\{t_{\text{AI}},t_{\text{DB}}\} &
\omega_8&=\{t_{\text{AI}},t_{\text{ML}},t_{\text{DB}}\} &
\omega_9&=\{t_{\text{AI}},t_{\text{KR}},t_{\text{DB}}\} &
\omega_{10}&=\{t_{\text{AI}},t_{\text{ML}},t_{\text{KR}},t_{\text{DB}}\} 
\end{align*}
Accordingly, we can represent nodes using 10-dimensional embeddings. The input embeddings of the topic nodes would then be given as:
\begin{align*}
\mathbf{t_{\text{AI}}^{(0)}} &= (0,-1,-1,-1,-1,0,-1,-1,-1,-1)\\
\mathbf{t_{\text{ML}}^{(0)}} &= (0,0,-1,0,-1,0,0,-1,0,-1)\\
\mathbf{t_{\text{KR}}^{(0)}} &= (0,0,0,-1,-1,0,0,0,-1,-1)\\
\mathbf{t_{\text{DB}}^{(0)}} &= (0,0,0,0,0,-1,-1,-1,-1,-1)
\end{align*}
Note, for instance, how $\max(\mathbf{t_{\text{AI}}^{(0)}},\mathbf{t_{\text{ML}}^{(0)}})=\mathbf{t_{\text{ML}}^{(0)}}$. As a result, an article node that is connected to both $t_{\text{AI}}$ and $t_{\text{ML}}$ would have the same embedding as an article node that is only connected to $t_{\text{ML}}$, which reflects the fact that the rule $t_{\text{ML}} \rightarrow t_{\text{AI}}$ has been taken into account as background knowledge.

\subsection{Non-monotonic reasoning}\label{secExNMR} Our focus on accumulating knowledge, and the use of the union in the formulation of the epistemic pooling principle, may suggest that the problem setting we consider in this paper is only suitable for monotonic reasoning. In contrast, many applications require some kind of non-monotonic inference, where we can defeasibly infer something to be true based on the absence of evidence to the contrary. To illustrate this idea, let us again consider the topics $t_{\text{AI}}$, $t_{\text{ML}}$, $t_{\text{KR}}$ and $t_{\text{DB}}$ and the background knowledge expressed in \eqref{eqExBackgroundKnowledge1}--\eqref{eqExBackgroundKnowledge2}. Let us now additionally assume that we want to implement the following behaviour:
\begin{itemize}
\item If we know that the topic of an article is AI, and we have no evidence that the article is about KR, then we will assume that the topic is ML (i.e.\ ``AI articles are typically about ML'').
\item However, if we know that the article is about AI and KR, then this inference should not be made.
\end{itemize}
This intuition cannot be implemented using standard propositional entailment, given the non-monotonic nature of the desired inferences: we cannot at the same time have $K\cup \{t_{\text{AI}}\} \models t_{\text{ML}}$ and $K\cup \{t_{\text{AI}},t_{\text{KR}}\} \not\models t_{\text{ML}}$ for any knowledge base $K$. The standard solution is to assume that each possible world $\omega$ has some plausibility degree $\pi(\omega)$ and to interpret ``if $\alpha$ then typically $\beta$'' as the constraint that $\max\{\pi(\omega) \,|\, \omega\models \alpha\wedge \beta\} > \max\{\pi(\omega) \,|\, \omega\models \alpha\wedge \neg \beta\}$ \cite{DBLP:journals/ai/LehmannM92,DBLP:conf/tark/Pearl90,DBLP:journals/ai/BenferhatDP97,DBLP:journals/jacm/FriedmanH01}. In other words, such a default rule is satisfied if $\beta$ is true in the most plausible models of $\alpha$. Using possibilistic logic, we can conveniently encode default knowledge using weighted logical formulas \cite{DBLP:journals/ai/BenferhatDP97}. In particular, the available knowledge can be encoded as follows:
\begin{align*}
1:      && t_{\text{ML}} &\rightarrow t_{\text{AI}} \\
1:      && t_{\text{KR}} &\rightarrow t_{\text{AI}} \\
0.5:   && t_{\text{KR}} &\rightarrow \bot \\
0.5:    && t_{\text{AI}} &\rightarrow t_{\text{ML}}
\end{align*}
Let $K_{\lambda}$ be the set of formulas whose weights is at least $\lambda$. To perform defeasible reasoning in possibilistic logic, we proceed as follows. Let the available evidence be encoded as a propositional formula $\alpha$. We first find the lowest $\lambda\geq 0$ for which $K_{\lambda}\cup \{\alpha\}$ is logically consistent. Then we say that $\beta$ is defeasibly entailed from the evidence $\alpha$ if it is classically entailed from $K_{\lambda}\cup \{\alpha\}$. It can be straightforwardly verified that the weighted formulas above capture the intended default knowledge. For instance, if $\alpha=t_{\text{AI}}$, we have $\lambda=0$. Since $K_0$ contains the rule $t_{\text{AI}} \rightarrow t_{\text{ML}}$ we can defeasibly infer $t_{\text{ML}}$. However, if the available evidence is $t_{\text{AI}}\wedge t_{\text{KR}}$, then we find $\lambda=1$ and we can no longer infer $t_{\text{ML}}$. Note how the rule $t_{\text{KR}} \rightarrow \bot$ intuitively serves the purpose of blocking the rule $t_{\text{AI}} \rightarrow t_{\text{ML}}$ in situations where $t_{\text{KR}}$ is known to be true.

Let $\omega_1,...,\omega_{10}$ be defined as before and let us consider, for this example, three plausibility levels. The main intuition is as follows: the higher the value of a given coordinate in the embeddings, the more strongly we can exclude the corresponding possible world. In particular, if the available evidence is violated in a possible world, then its corresponding coordinate is set to 1. Otherwise, if a formula with a weight of 0.5 is violated, the corresponding coordinate is set to 0.5. Note that the formulas of weight 1 are  taken into account implicitly, as we only consider the models of these formulas as our 10 possible worlds. The remaining coordinates are set to 0. The initial embeddings of the topic nodes are then defined as follows:
\begin{align*}
\mathbf{t_{\text{AI}}^{(0)}} &= (1,0.5,0,0.5,0.5,1,0.5,0,0.5,0.5)\\
\mathbf{t_{\text{ML}}^{(0)}} &= (1,1,0,1,0.5,1,1,0,1,0.5)\\
\mathbf{t_{\text{KR}}^{(0)}} &= (1,1,1,0.5,0.5,1,1,1,0.5,0.5)\\
\mathbf{t_{\text{DB}}^{(0)}} &= (1,1,1,1,1,0,0.5,0,0.5,0.5)
\end{align*}
To test whether a propositional formula $\alpha$ can be defeasibly inferred in the epistemic state encoded by a vector $\mathbf{x}=(x_1,...,x_{n})$, we need to check whether $\min\{x_i \,|\, \omega_i\models \alpha\} < \min\{x_i \,|\, \omega_i\models \neg\alpha\}$. Indeed, if this inequality holds, there is a model of $\alpha$ which is strictly more plausible than any of the models of $\neg \alpha$. For instance, going back to the example, let $a_1$ be an article node which is only connected to the topic node $t_{\text{AI}}$. The embedding of this article $\mathbf{a_1^{(1)}}$ is given by $(1,0.5,0,0.5,0.5,1,0.5,0,0.5,0.5)$. Note that $t_{\text{ML}}$ is satisfied in worlds $\omega_3,\omega_5,\omega_8,\omega_{10}$. We find that:
\begin{align*}
\min\{x_i \,|\, \omega_i\models t_{\text{ML}}\}
&=\min(x_3,x_5,x_8,x_{10}) = 0\\
\min\{x_i \,|\, \omega_i\models \neg t_{\text{ML}}\}
&=\min(x_1,x_2,x_4,x_6,x_7,x_9) = 0.5
\end{align*}
Hence we can indeed defeasibly infer that the article is about Machine Learning. Let $a_2$ be an article which connected to both $t_{\text{AI}}$ and $t_{\text{KR}}$. The embedding $\mathbf{a_2^{(1)}}$ is then given by 
\begin{align*}
\mathbf{a_2^{(1)}}&=\max((1,0.5,0,0.5,0.5,1,0.5,0,0.5,0.5),(1,1,1,0.5,0.5,1,1,1,0.5,0.5))\\
&= (1,1,1,0.5,0.5,1,1,1,0.5,0.5)
\end{align*}
We then find:
\begin{align*}
\min\{x_i \,|\, \omega_i\models t_{\text{ML}}\}
&=\min(x_3,x_5,x_8,x_{10}) = 0.5\\
\min\{x_i \,|\, \omega_i\models \neg t_{\text{ML}}\}
&=\min(x_1,x_2,x_4,x_6,x_7,x_9) = 0.5
\end{align*}
It is thus not possible to defeasibly infer that $a_2$ is about Machine Learning.
As this example illustrates, non-monotonic reasoning essentially requires that we can model weighted epistemic states. This will be studied in Section \ref{secWeightedEpistemicStates}, where we will see that max-pooling is uniquely suitable for this setting.

\section{Problem setting}\label{secProblemSetting}
In this section, we introduce our considered problem setting more formally.
We assume that epistemic states are represented using sets of elementary properties. Let us write $\mathcal{P}$ for the set of all these properties. An epistemic state $\mathcal{Q}$ then simply corresponds to a subset of $\mathcal{P}$. Intuitively, we think of these elementary properties as atomic pieces of evidence. The properties in an epistemic state $\mathcal{Q}$ then correspond to the evidence that is available, whereas the properties in $\mathcal{P}\setminus \mathcal{Q}$ correspond to evidence that has not been encountered. For instance, in the context of image processing, we can think of the properties from $\mathcal{P}$ as elementary visual features, whose presence may be detected in an image. In applications where more intricate forms of reasoning are needed than simply aggregating sets of detected features, we can relate the properties in $\mathcal{P}$ to possible worlds, as we have seen in Section \ref{secExPropositional}. For each possible world $\omega$, we then consider a property $p_{\omega}$, corresponding to the knowledge that $\omega$ can be excluded, i.e.\ that $\omega$ is not a model of the world. In this way, subsets of $\mathcal{P}$ can be used to represent arbitrary propositional knowledge bases. This link with logical reasoning will be developed in Section \ref{secPropositionalReasoning}. For now, however, it will suffice to simply think of epistemic states as subsets of $\mathcal{P}$.

Taking the view that embeddings encode epistemic states, each $\mathbf{e}\in\mathbb{R}^n$ will be associated with a set of properties from $\mathcal{P}$.  Formally, we assume that a scoring function $\gamma_p:\mathbb{R}^n\rightarrow \mathbb{R}$ is available for each property $p\in \mathcal{P}$. We consider two variants of our setting, which differ in whether strict or weak inequalities are used to determine which properties are satisfied. As we will see, this choice has a material impact on the theoretical properties of the resulting framework.

\paragraph{Strict semantics}
Under the strict semantics, we say that an embedding $\mathbf{e}\in\mathbb{R}^n$ satisfies the property $p\in \mathcal{P}$ if $\gamma_p(\mathbf{e})>0$. 
Let us write $\Gamma(\mathbf{e})$ for the epistemic state encoded by $\mathbf{e}$, i.e.\ the set of properties satisfied by $\mathbf{e}$:
\begin{align}\label{eqDefGamma}
\Gamma(\mathbf{e}) = \{p\in \mathcal{P} \mid \gamma_p(\mathbf{e})>0\}
\end{align}
Let $\diamond: \mathbb{R}^n \times \mathbb{R}^n \rightarrow \mathbb{R}^n$ represent a pooling operator. The epistemic pooling principle can then be formalised as follows:
\begin{align}\label{eqEpistemicPoolingGeneral}
\Gamma(\mathbf{e} \diamond \mathbf{f}) = \Gamma(\mathbf{e}) \cup \Gamma(\mathbf{f})
\end{align}
If we want to specify that the strict semantics is used, we will also refer to \eqref{eqEpistemicPoolingGeneral} as the 
\emph{strict epistemic pooling principle}.
Intuitively, the embeddings $\mathbf{e}$ and $\mathbf{f}$ capture information coming from two different sources, e.g.\ two different regions of an image or two different modalities. The principle captured by \eqref{eqEpistemicPoolingGeneral} is that the pooling operator $\diamond$ should merely combine this information: the total evidence that is available is the union of the evidence provided by the two sources. Note that \eqref{eqEpistemicPoolingGeneral} is equivalent to:
\begin{align}
\forall p\in\mathcal{P}. (\gamma_{p}(\mathbf{e})>0) \vee (\gamma_{p}(\mathbf{f})>0) \Leftrightarrow \gamma_{p}(\mathbf{e}  \diamond \mathbf{f}) > 0
\end{align}
In the following, we will assume that all embeddings are taken from some set $X\subseteq \mathbb{R}^n$. One possibility would be to choose $X=\mathbb{R}^n$, but as we will see, it is sometimes necessary to make a more restrictive choice. For instance, we may have $X=[0,+\infty[^n$ if we want to restrict the discussion to vectors with non-negative coordinates. Regardless of how $X$ is chosen, an important consideration is that the embeddings in $X$ should allow us to capture every possible epistemic state, in the following sense:
\begin{align}\label{eqAvgAssumptionEveryCombination}
\forall \mathcal{Q}\subseteq \mathcal{P}. \exists \mathbf{e}\in X.\Gamma(\mathbf{e})=\mathcal{Q}
\end{align}
If \eqref{eqAvgAssumptionEveryCombination} is satisfied, we say that $X$ satisfies \emph{exhaustiveness}.
Finally, for the ease of presentation, we introduce the following notations:
\begin{align*}
\mathsf{Pos}_p &= \{\mathbf{e} \in X \mid \gamma_p(\mathbf{e})>0\} \\
\mathsf{Neg}_p &= \{\mathbf{e} \in X \mid \gamma_p(\mathbf{e})\leq 0\} = X\setminus \mathsf{Pos}_p
\end{align*}
We will refer to $\mathsf{Pos}_p$ and $\mathsf{Neg}_p$ as the positive and negative regions for property $p$. Indeed, we have that $\mathbf{e}\in \mathsf{Pos}_p$ iff $p\in \Gamma(\mathbf{e})$, and $\mathbf{e}\in \mathsf{Neg}_p$ otherwise.

\paragraph{Weak epistemic pooling principle}
Under the weak semantics, we say that an embedding $\mathbf{e}\in\mathbb{R}^n$ satisfies the property $p\in \mathcal{P}$ if $\gamma_p(\mathbf{e})\geq 0$. 
Epistemic states are then determined as follows:
\begin{align}\label{eqDefGammaPrime}
\Gamma'(\mathbf{e}) = \{p\in \mathcal{P} \mid \gamma_p(\mathbf{e})\geq 0\}
\end{align}
This definition gives rise to the following counterpart of \eqref{eqEpistemicPoolingGeneral}, which we will refer to as the \emph{weak epistemic pooling principle}: 
\begin{align}\label{eqEpistemicPoolingWeak}
\Gamma'(\mathbf{e} \diamond \mathbf{f}) = \Gamma'(\mathbf{e}) \cup \Gamma'(\mathbf{f})
\end{align}
We will furthermore require that \emph{exhaustiveness} is satisfied:
\begin{align}\label{eqAvgAssumptionEveryCombinationWeak}
\forall \mathcal{Q}\subseteq \mathcal{P}. \exists \mathbf{e}\in X.\Gamma'(\mathbf{e})=\mathcal{Q}
\end{align}
Finally, the positive and negative regions are also defined analogously as before:
\begin{align*}
\mathsf{Pos}'_p &= \{\mathbf{e} \in X \mid \gamma_p(\mathbf{e})\geq 0\} \\
\mathsf{Neg}'_p &= \{\mathbf{e} \in X \mid \gamma_p(\mathbf{e})< 0\} = X \setminus \mathsf{Pos}'_p
\end{align*}

\paragraph{Pooling operators}
Whether the (strict or weak) epistemic pooling principle can be satisfied for every $\mathbf{e},\mathbf{f}\in X$ depends on the choice of the scoring functions $\gamma_p$, the set $X\subseteq \mathbb{R}^n$ and the pooling operator $\diamond$. In our analysis, we will focus on the following standard pooling operators:
\begin{align*}
&\textbf{Average:} & (e_1,...,e_n) \diamond_{\mathsf{avg}} (f_1,...,f_n) &= \frac{\mathbf{e} + \mathbf{f}}{2}\\
&\textbf{Summation:} & (e_1,...,e_n) \diamond_{\mathsf{sum}} (f_1,...,f_n) &= \mathbf{e} + \mathbf{f}\\
&\textbf{Max-pooling:} & (e_1,...,e_n) \diamond_{\mathsf{max}} (f_1,...,f_n) &= (\max(e_1,f_1),...,\max(e_n,f_n))\\
&\textbf{Hadamard:} & (e_1,...,e_n) \diamond_{\mathsf{had}} (f_1,...,f_n) &= (e_1\cdot f_1,...,e_n \cdot f_n)
\end{align*}
Note that the epistemic pooling principles in \eqref{eqEpistemicPoolingGeneral} and \eqref{eqEpistemicPoolingWeak} are defined w.r.t.\ two arguments. This focus on binary pooling operators simplifies the formulation, while any negative results we obtain naturally carry over to pooling operators with more arguments. Moreover, most of the considered pooling operators are associative, with the exception of $\diamond_{\textit{avg}}$. 
Furthermore, even though $\diamond_{\textit{avg}}$ itself is not associative, if it satisfies \eqref{eqEpistemicPoolingGeneral} or \eqref{eqEpistemicPoolingWeak}, its effect on the epistemic states encoded by the embeddings will nonetheless be associative, given that we have e.g.\ $\Gamma(\mathbf{e_1} \diamond (\mathbf{e_2}\diamond \mathbf{e_3})) = \Gamma((\mathbf{e_1} \diamond \mathbf{e_2}) \diamond \mathbf{e_3})= \Gamma(\mathbf{e_1})\cup \Gamma(\mathbf{e_2}) \cup \Gamma(\mathbf{e_3})$, due to the associativity of the union. We now illustrate the key concepts with a simple example.
\begin{example}
Let $\mathcal{P}=\{a,b\}$ and suppose embedding are taken from $\mathbb{R}^2$. Let the scoring functions $\gamma_a$ and $\gamma_b$ be defined as follows:
\begin{align*}
\gamma_a(x_1,x_2) &= 1 - d((x_1,x_2),(0,0))  = 1- \sqrt{x_1^2 + x_2^2} \\
\gamma_b(x_1,x_2) &= 1 - d((x_1,x_2),(1,1))  = 1- \sqrt{(1-x_1)^2 + (1-x_2)^2} 
\end{align*}
Now let $\mathbf{e}=(\frac{1}{4},0)$ and $\mathbf{f}=(\frac{3}{4},1)$. Then we have $\mathbf{e}\diamond_{\mathsf{avg}}\mathbf{f}=(\frac{1}{2},\frac{1}{2})$. We find:
\begin{align*}
\gamma_a(\mathbf{e}) &=  \frac{3}{4}&
\gamma_a(\mathbf{f}) &= -\frac{1}{4}&
\gamma_a(\mathbf{e}\diamond_{\mathsf{avg}}\mathbf{f}) &=1-\frac{1}{\sqrt{2}}\\
\gamma_b(\mathbf{e}) &=  -\frac{1}{4}&
\gamma_b(\mathbf{f}) &= \frac{3}{4}&
\gamma_b(\mathbf{e}\diamond_{\mathsf{avg}}\mathbf{f}) &=1-\frac{1}{\sqrt{2}}
\end{align*}
and thus
\begin{align*}
\Gamma(\mathbf{e}) &= \{a\} &
\Gamma(\mathbf{f}) &= \{b\} &
\Gamma(\mathbf{e}\diamond_{\mathsf{avg}}\mathbf{f}) &= \{a,b\}
\end{align*}
This means that the epistemic pooling principle \eqref{eqEpistemicPoolingGeneral} is satisfied for $\mathbf{e}$ and $\mathbf{f}$. On the other hand, for $\mathbf{g} = (10,10)$, we have $\Gamma(\mathbf{g})=\Gamma(\mathbf{e}\diamond_{\mathsf{avg}}\mathbf{g})=\emptyset$, hence the epistemic pooling principle is not satisfied for $\mathbf{e}$ and $\mathbf{g}$.
\end{example}

\paragraph{Notations}
Throughout this paper, we write $\delta(A)$ for the boundary of a set $A\subseteq \mathbb{R}^n$. Similarly, we will write $\textit{int}(A)$ and $\textit{cl}(A)$ for the interior and closure:
\begin{align*}
\textit{int}(A) &= \{\mathbf{e} \in A \mid \exists \varepsilon >0 \,.\, \forall \mathbf{f}\in \mathbb{R}^n\,.\, d(\mathbf{e},\mathbf{f})<\varepsilon \Rightarrow \mathbf{f}\in A\}\\
\textit{cl}(A) &=\{\mathbf{e} \in \mathbb{R}^n \mid \forall \varepsilon >0\,.\, \exists \mathbf{f}\in A\,.\, d(\mathbf{e},\mathbf{f})<\varepsilon\}\\
\delta(A)&=\textit{cl}(A)\setminus \textit{int}(A)
\end{align*}
\section{Realizability of the epistemic pooling principle}\label{secPoolingProperties}
In this section we study, for each of the considered pooling operators, whether they can satisfy the strict and weak epistemic pooling principles, and if so, under which conditions this is the case. In all cases, we find that the epistemic pooling principles can only be satisfied if $n\geq |\mathcal{P}|$, with $n$ the dimensionality of the embeddings. For $\diamond_{\mathsf{avg}}$ and $\diamond_{\mathsf{sum}}$, we also have to make assumptions on the set $X$, i.e.\ the epistemic pooling principles cannot be satisfied for $X=\mathbb{R}^n$ with these pooling operators. We furthermore find that the epistemic pooling principles can only be satisfied if $\gamma_p$ satisfies some particular conditions. Most significantly, we find that $\diamond_{\mathsf{avg}}$ and $\diamond_{\mathsf{sum}}$ cannot satisfy the weak epistemic pooling principle with continuous scoring functions $\gamma_p$, and that $\diamond_{\mathsf{had}}$ cannot satisfy the strict epistemic pooling principle with continuous scoring functions. These results are summarised in Table \ref{tabConditionsEpistemicPooling}.

\begin{table}
\centering
\begin{tabular}{lccc}
\toprule
\textbf{Pooling Operator}  & \textbf{Semantics} & \textbf{$X=\mathbb{R}^n$ possible?} & \textbf{Continuous $\gamma_p$ possible?}\\
\midrule
\multirow{2}{*}{Average} & Strict & \xmark & \cmark\\
        & Weak & \xmark & \xmark\\
\midrule
\multirow{2}{*}{Summation} & Strict & \xmark & \cmark\\
        & Weak & \xmark  & \xmark\\
\midrule
\multirow{2}{*}{Max-pooling} & Strict & \cmark & \cmark\\
 & Weak & \cmark & \cmark\\
\midrule
\multirow{2}{*}{Hadamard}  & Strict & \cmark & \xmark\\
        & Weak & \cmark & \cmark\\
\bottomrule
\end{tabular}
\caption{Summary of results about the realizability of the epistemic pooling principles. \label{tabConditionsEpistemicPooling}}
\end{table}

\subsection{Average}\label{secRealizationAverage}
\paragraph{Strict semantics}
The first question we look at is whether the strict epistemic pooling principle \eqref{eqEpistemicPoolingGeneral} can be satisfied for all $\mathbf{e},\mathbf{f}\in \mathbb{R}^n$. The following result shows that this is only possible in the trivial case where every embedding encodes the same epistemic state.
\begin{proposition}\label{propAvgWeakNeedsX}
Suppose the epistemic pooling principle \eqref{eqEpistemicPoolingGeneral} is satisfied for all embeddings $\mathbf{e},\mathbf{f}\in \mathbb{R}^n$, with $\diamond=\diamond_{\mathsf{avg}}$. For any given $p\in \mathcal{P}$ we have
$$
(\forall \mathbf{e}\in\mathbb{R}^n\,.\, p\in \Gamma(\mathbf{e})) \vee (\forall \mathbf{e}\in\mathbb{R}^n\,.\, p\notin \Gamma(\mathbf{e}))
$$
\end{proposition}
\begin{proof}
Suppose there exists some $\mathbf{e}\in\mathbb{R}^n$ such that $p\in \Gamma(\mathbf{e})$. We show that we then have $p\in \Gamma(\mathbf{f})$ for every $\mathbf{f}\in\mathbb{R}^n$. Noting that $\mathbf{f} = \mathbf{e}\diamond_{\mathsf{avg}} (2\mathbf{f} - \mathbf{e})$, we know from \eqref{eqEpistemicPoolingGeneral} that $\Gamma(\mathbf{f}) = \Gamma(\mathbf{e}) \cup \Gamma(2\mathbf{f} - \mathbf{e})$, and thus in particular that $p\in \Gamma(\mathbf{f})$.
\end{proof}
We will thus have to assume that embeddings are restricted to some subset $X\subset \mathbb{R}^n$. To ensure that $X$ is closed under the pooling operator $\diamond_{\mathsf{avg}}$, we will assume that $X$ is convex. The following three lemmas explain how \eqref{eqEpistemicPoolingGeneral} constrains the scoring functions $\gamma_p$. 
\begin{lemma}\label{lemmaAvgBoundary}
Suppose the epistemic pooling principle  \eqref{eqEpistemicPoolingGeneral} is satisfied for all embeddings $\mathbf{e},\mathbf{f}\in X$, with $\diamond=\diamond_{\mathsf{avg}}$ and $X$ convex. Suppose there exists some $\mathbf{e}\in X$ such that $p\in \Gamma(\mathbf{e})$.
It holds that $\mathsf{Neg}_p \subseteq \delta(X)$.
\end{lemma}
\begin{proof}
Suppose $\mathbf{f}\in \textit{int}(X)$. We show that $\mathbf{f}\in \mathsf{Pos}_{p}$. Let us define ($\lambda\in \mathbb{R}$):
$$
\mathbf{x}_{\lambda} = \lambda\mathbf{e} + (1-\lambda)\mathbf{f}
$$
Note that because we assumed that $X$ is convex, it holds that $\mathbf{x}_{\lambda}\in X$ for all $\lambda\in[0,1]$.
We have $p\in \Gamma(\mathbf{x}_1)$, as $\mathbf{x}_1=\mathbf{e}$. By repeatedly applying \eqref{eqEpistemicPoolingGeneral} we find that $p$ belongs to $\Gamma(\mathbf{x}_{\frac{1}{2}})$, $\Gamma(\mathbf{x}_{\frac{1}{4}})$, $\Gamma(\mathbf{x}_{\frac{1}{8}})$, etc. In the limit, we find that for every $\lambda\in ]0,1]$ it holds that $p\in \Gamma(\mathbf{x}_\lambda)$. Since $\mathbf{f}\in \textit{int}(X)$, there exists some $0<\varepsilon\leq 1$ such that $\mathbf{x}_{-\varepsilon}\in X$. Using \eqref{eqEpistemicPoolingGeneral}, we find $\Gamma(\mathbf{f}) = \Gamma(\mathbf{x}_{-\varepsilon}) \cup \Gamma(\mathbf{x}_{\varepsilon})$. Since we established $p\in \Gamma(\mathbf{x}_{\varepsilon})$ we thus find $p\in \Gamma(\mathbf{f})$, which means $\mathbf{f}\in \mathsf{Pos}_{p}$.
\end{proof}

\begin{corollary}
Suppose the epistemic pooling principle  \eqref{eqEpistemicPoolingGeneral} is satisfied for all embeddings $\mathbf{e},\mathbf{f}\in X$, with $\diamond=\diamond_{\mathsf{avg}}$ and $X$ convex. Let $p\in \mathcal{P}$. It holds that $\textit{dim}(\mathsf{Neg}_p)\leq n-1$
\end{corollary}

\begin{lemma}\label{lemmaAvgConvex1}
Suppose the epistemic pooling principle  \eqref{eqEpistemicPoolingGeneral} is satisfied for all embeddings $\mathbf{e},\mathbf{f}\in X$, with $\diamond=\diamond_{\mathsf{avg}}$ and $X$ convex. Let $p\in \mathcal{P}$. It holds that $\mathsf{Pos}_p$ is convex.
\end{lemma}
\begin{proof}
Suppose $\mathbf{e},\mathbf{f}\in \mathsf{Pos}_p$ and define ($\lambda\in[0,1]$):
$$
\mathbf{x}_{\lambda} = \lambda\mathbf{e} + (1-\lambda)\mathbf{f}
$$ 
We show that $\mathbf{x}_{\lambda}\in \mathsf{Pos}_p$ for every $\lambda\in]0,1[$.
By applying \eqref{eqEpistemicPoolingGeneral} to $\mathbf{f}=\mathbf{x}_0$ and $\mathbf{e}=\mathbf{x}_1$ we find that $\mathbf{x}_{\frac{1}{2}}\in \mathsf{Pos}_p$. By applying \eqref{eqEpistemicPoolingGeneral} to $\mathbf{x}_0$ and $\mathbf{x}_{\frac{1}{2}}$, we find $\mathbf{x}_{\frac{1}{4}}\in \mathsf{Pos}_p$. Similarly, by applying \eqref{eqEpistemicPoolingGeneral} to $\mathbf{x}_{\frac{1}{2}}$ and $\mathbf{x}_{1}$, we find $\mathbf{x}_{\frac{3}{4}}\in \mathsf{Pos}_p$. Continuing in this way, we find $\mathbf{x}_{\lambda} \in \mathsf{Pos}_p$ for every $\lambda$ of the form $\frac{j}{2^i}$ with $i\in\mathbb{N}$ and $j\in\{0,1,...,2^i\}$. Now let $\lambda\in ]0,1[$. We can approximate $\lambda$ arbitrary well using a value of the form $\frac{j}{2^i}$. In particular, we can always find some $i\in \mathbb{N}$ and $j\in\{0,...,2^i\}$ such that $0<\frac{j}{2^i}<\lambda$ and $\lambda < 2\lambda- \frac{j}{2^i}<1$. By \eqref{eqEpistemicPoolingGeneral}, we have:
$$
\Gamma(\mathbf{x}_{\lambda})=\Gamma\left(\mathbf{x}_{\frac{j}{2^i}}\right)\cup \Gamma\left(\mathbf{x}_{2\lambda- \frac{j}{2^i}}\right)
$$
Since we already know that $\mathbf{x}_{\frac{j}{2^i}} \in \mathsf{Pos}_p$ we thus also find $\mathbf{x}_{\lambda}\in\mathsf{Pos}_p$.
\end{proof}

\begin{figure}
\centering
\includegraphics[width=175pt]{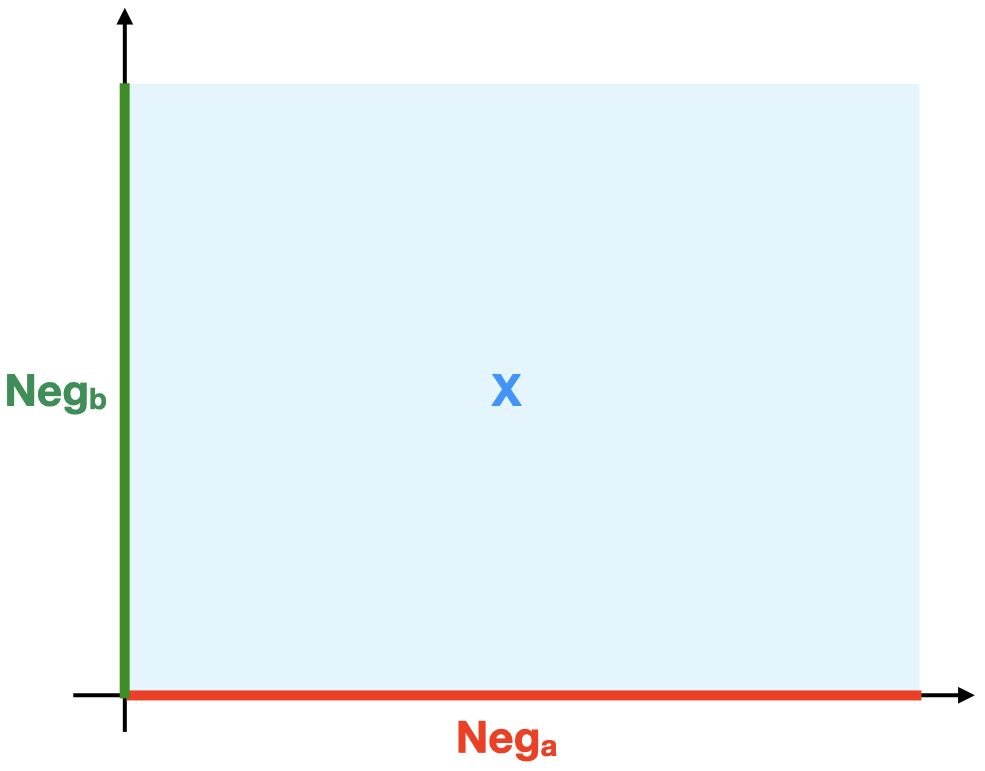}
\caption{Construction satisfying \eqref{eqEpistemicPoolingGeneral} and \eqref{eqAvgAssumptionEveryCombination} \label{figAvgPoolingConstruction} for $\diamond_{\mathsf{avg}}$.}
\end{figure}

\begin{lemma}\label{lemmaAvgConvex2}
Suppose the epistemic pooling principle  \eqref{eqEpistemicPoolingGeneral} is satisfied for all embeddings $\mathbf{e},\mathbf{f}\in X$, with $\diamond=\diamond_{\mathsf{avg}}$ and $X$ convex. Let $p\in \mathcal{P}$. It holds that $\mathsf{Neg}_p$ is convex.
\end{lemma}
\begin{proof}
Suppose $\mathbf{e},\mathbf{f}\in \mathsf{Neg}_p$ and define ($\lambda\in[0,1]$):
$$
\mathbf{x}_{\lambda} = \lambda\mathbf{e} + (1-\lambda)\mathbf{f}
$$ 
Entirely analogously as in the proof of Lemma \ref{lemmaAvgConvex1}, we find that $\mathbf{x}_{\lambda}\in \mathsf{Neg}_p$ for every $\lambda$ of the form $\frac{j}{2^i}$, with $i\in \mathbb{N}$ and $j\in\{0,...,2^i\}$. Let $\lambda \in ]0,1[$. Let us consider values $\varepsilon_1,\varepsilon_2$ satisfying $0<\varepsilon_1 < \varepsilon_2 < \lambda < 2_\lambda-\varepsilon_2 < 2_\lambda-\varepsilon_1 <1$. 
Suppose $\mathbf{x}_{\lambda}\notin \mathsf{Neg}_p$. Then from \eqref{eqEpistemicPoolingGeneral} we know that:
\begin{itemize}
\item  $\mathbf{x}_{\varepsilon_1}\in \mathsf{Pos}_p$ or $\mathbf{x}_{2\lambda -\varepsilon_1}\in \mathsf{Pos}_p$ needs to hold; and
\item $\mathbf{x}_{\varepsilon_2}\in \mathsf{Pos}_p$ or $\mathbf{x}_{2\lambda -\varepsilon_2}\in \mathsf{Pos}_p$ needs to hold.
\end{itemize}
This means that there are at least two distinct values $\lambda_1,\lambda_2 \in \{\varepsilon_1,\varepsilon_2,2\lambda-\varepsilon_1,2\lambda-\varepsilon_2\}$ such that $\mathbf{x}_{\lambda_1}\in \mathsf{Pos}_p$ and $\mathbf{x}_{\lambda_2}\in \mathsf{Pos}_p$. Let us assume w.l.o.g.\ that $\lambda_1<\lambda_2$. We can always find some $i\in\mathbb{N}$ and $j\in\{0,...,2^i\}$ such that $\lambda_1<\frac{j}{2^i}<\lambda_2$. From the preceding discussion we already know that $\mathbf{x}_{\frac{j}{2^i}}\in \mathsf{Neg}_p$. However, from $\mathbf{x}_{\lambda_1}\in \mathsf{Pos}_p$ and $\mathbf{x}_{\lambda_2}\in \mathsf{Pos}_p$, using Lemma \ref{lemmaAvgConvex1} we find $\mathbf{x}_{\frac{j}{2^i}}\in \mathsf{Pos}_p$, a contradiction. It follows that $\mathbf{x}_{\lambda}\in \mathsf{Neg}_p$
\end{proof}
%
%
%
From Lemmas \ref{lemmaAvgBoundary}, \ref{lemmaAvgConvex1} and \ref{lemmaAvgConvex2}, we obtain the following corollary using the hyperplane separation theorem.
\begin{corollary}
Suppose the epistemic pooling principle  \eqref{eqEpistemicPoolingGeneral} is satisfied for all embeddings $\mathbf{e},\mathbf{f}\in X$, with $\diamond=\diamond_{\mathsf{avg}}$ and $X$ convex. 
For any $p\in \mathcal{P}$, there exists a hyperplane $H_p$ such that  $\mathsf{Neg}_p\subseteq \delta(X)\cap H_p$.
\end{corollary}
The next proposition reveals that the dimensionality of the embeddings needs to be at least $\vert\mathcal{P}\vert$ if we want the epistemic pooling principle to be satisfied and at the same time ensure that every epistemic state is modelled by some vector.
\begin{proposition}\label{propDimAvg1}
Suppose the epistemic pooling principle  \eqref{eqEpistemicPoolingGeneral} is satisfied for all embeddings $\mathbf{e},\mathbf{f}\in X$, with $\diamond=\diamond_{\mathsf{avg}}$ and $X\subseteq \mathbb{R}^n$ convex. Suppose furthermore that exhaustiveness \eqref{eqAvgAssumptionEveryCombination} is satisfied. It holds that $n\geq \vert\mathcal{P}\vert$.
\end{proposition}
\begin{proof}
Let $p_1,...,p_{\vert\mathcal{P}\vert}$ be an enumeration of the properties in $\mathcal{P}$. 
Note that because of \eqref{eqAvgAssumptionEveryCombination}, we have that $\mathsf{Neg}_{p_1}\neq \emptyset$ and $\mathsf{Pos}_{p_1}\neq\emptyset$. Moreover, from Lemmas \ref{lemmaAvgConvex1} and \ref{lemmaAvgConvex2} we know that these regions are both convex. It follows from the hyperplane separation theorem that there exists a hyperplane $H_1$ which separates $\mathsf{Neg}_{p_1}$ and $\mathsf{Pos}_{p_1}$. From Lemma \ref{lemmaAvgBoundary}, we furthermore know that $\mathsf{Neg}_{p_1}\subseteq \textit{cl}(\mathsf{Pos}_{p_1})$, which implies that $\mathsf{Neg}_{p_1}\subseteq H_1$.

Note that $H_1\cap \mathsf{Neg}_{p_2}$ and $H_1\cap \mathsf{Pos}_{p_2}$ are convex regions. Moreover, since $\mathsf{Neg}_{p_1}\subseteq H_1$, we find from \eqref{eqAvgAssumptionEveryCombination} that $H_1\cap \mathsf{Neg}_{p_2}\neq \emptyset$ and $H_1\cap \mathsf{Pos}_{p_2}\neq\emptyset$. It follows from the hyperplane separation theorem that there exists some hyperplane $H_2$ separating $H_1\cap \mathsf{Neg}_{p_2}$ and $H_1\cap \mathsf{Pos}_{p_2}$. Moreover, it holds that $H_1\cap \mathsf{Neg}_{p_2} \subseteq H_2$. Indeed, suppose there was some $\mathbf{e} \in (H_1\cap \mathsf{Neg}_{p_2})\setminus H_2$ and let $\mathbf{f}\in H_1\cap \mathsf{Pos}_{p_2}$. For $i\in \mathbb{N}\setminus \{0\}$ we define $\mathbf{f}_i= \mathbf{e}\diamond_{\mathsf{avg}}\mathbf{f}_{i-1}$, with $\mathbf{f}_{0}= \mathsf{f}$. Then there must be some $i\in \mathbb{N}\setminus \{0\}$ such that $\mathbf{f}_i$ is on the same side of hyperplane $H_2$ as $\mathbf{e}$, which implies $\mathbf{f}_i\in \mathsf{Neg}_{p_2}$ since $H_2$ was chosen as a separating hyperplane. However, using \eqref{eqEpistemicPoolingGeneral} we also find that $p_2\in \Gamma(\mathbf{f}_i)$ and thus $\mathbf{f}_i\in \mathsf{Pos}_{p_2}$, which is a contradiction. This means that $H_1\cap \mathsf{Neg}_{p_2} \subseteq H_2$ and thus in particular also that $\mathsf{Neg}_{p_1} \cap \mathsf{Neg}_{p_2} \subseteq H_2$.

We can repeat this argument as follows. For each $j\in \{1,...,\vert \mathcal{P}\vert\}$, we let $H_j$ be a hyperplane separating $\mathsf{Pos}_{p_j}$ and $\mathsf{Neg}_{p_j}$.
Let $i\in \{2,...,\vert\mathcal{P}\vert-1\}$ and suppose we have already shown for every $j\leq i$ that $H_1 \cap ... \cap H_{j-1} \cap \mathsf{Neg}_{p_j} \subseteq H_j$. Note that this in particular implies $\mathsf{Neg}_{p_1} \cap... \cap \mathsf{Neg}_{p_i} \subseteq H_i$. We know from \eqref{eqAvgAssumptionEveryCombination} that $\mathsf{Neg}_{p_1} \cap... \cap \mathsf{Neg}_{p_i} \cap \mathsf{Neg}_{p_{i+1}}\neq \emptyset$ and $\mathsf{Neg}_{p_1} \cap... \cap \mathsf{Neg}_{p_i} \cap \mathsf{Pos}_{p_{i+1}}\neq \emptyset$. It follows that $H_1 \cap ... \cap H_i \cap \mathsf{Neg}_{p_{i+1}}\neq \emptyset$ and $H_1 \cap ... \cap H_i \cap \mathsf{Pos}_{p_{i+1}}\neq \emptyset$. This implies that $H_1\cap ... \cap H_i \not\subseteq H_{i+1}$ and thus that $\textit{dim}(H_1\cap...
\cap H_{i+1}) = \textit{dim}(H_1\cap...
\cap H_{i})-1$. Similar as for the base case above, we furthermore find that $H_1 \cap ... \cap H_{i} \cap \mathsf{Neg}_{p_{i+1}} \subseteq H_{i+1}$.

It follows that $\textit{dim}(H_1\cap...
\cap H_{\vert\mathcal{P}\vert})=n- \vert\mathcal{P}\vert$, which is only possible if $n\geq \vert\mathcal{P}\vert$. 
\end{proof}
It is easy to see that the bound from this proposition cannot be strengthened, i.e.\ that it is possible to satisfy  \eqref{eqEpistemicPoolingGeneral} and \eqref{eqAvgAssumptionEveryCombination} while $n= \vert\mathcal{P}\vert$. One possible construction is as follows. Let $p_1,...,p_n$ be an enumeration of the properties in $\mathcal{P}$. We define:
$$
X = \{(x_1,...,x_n) \mid x_1,...,x_n\geq 0\}
$$
and for $i\in \{1,...,n\}$ we define:
$$
\gamma_{p_i}(e_1,...,e_n) = e_i
$$
To see why this choice satisfies \eqref{eqAvgAssumptionEveryCombination}, let $\mathcal{Q}\subseteq \mathcal{P}$ and define $\mathbf{e}=(e_1,...,e_n)$ as follows:
\begin{align*}
e_i = 
\begin{cases}
1 & \text{if $p_i\in \mathcal{Q}$}\\
0 & \text{otherwise}
\end{cases}
\end{align*}
Then it is straightforward to verify that $\Gamma(\mathbf{e})=\mathcal{Q}$. Moreover, it is also clear that \eqref{eqEpistemicPoolingGeneral} is satisfied. 
Indeed, the $i\textsuperscript{th}$ coordinate of $(e_1,...,e_n)\diamond_{\mathsf{avg}}(f_1,...,f_n)$ is $\frac{e_i+f_i}{2}$. Hence we have $\frac{e_i+f_i}{2}>0$ as soon as $e_i>0$ or $f_i>0$ (noting that $e_i,f_i\geq 0$), meaning $p_i\in \Gamma((e_1,...,e_n)\diamond_{\mathsf{avg}}(f_1,...,f_n))$ iff $p_i\in \Gamma(e_1,...,e_n)\cup \Gamma(f_1,...,f_n)$. A visualisation of this construction is shown in Figure \ref{figAvgPoolingConstruction}. Note how the construction aligns with the common practice of learning sparse high-dimensional embeddings with non-negative coordinates. 

\paragraph{Weak semantics}
Let us now consider whether the weak epistemic pooling principle can also be satisfied for $\diamond_{\mathsf{avg}}$. Without any restrictions on the scoring functions $\gamma_p$, this is clearly the case. In particular, suppose for each $p\in \mathcal{P}$, a function $\gamma_p$ is defined such that the strict epistemic pooling principle
\eqref{eqEpistemicPoolingGeneral} is satisfied for all embeddings $\mathbf{e},\mathbf{f}\in X$, for some convex set $X$, and suppose furthermore that exhaustiveness \eqref{eqAvgAssumptionEveryCombination} is satisfied. Then we can define modified scoring functions as follows:
\begin{align}\label{eqAvgWeakStrictDuality1}
\gamma'_p(\mathbf{e}) &=
\begin{cases}
1 & \text{if $\gamma_p(\mathbf{e}) >0$}\\
-1 & \text{otherwise}
\end{cases}
\end{align}
In particular, we have $\gamma_p(\mathbf{e})>0$ iff $\gamma'_p(\mathbf{e})\geq 0$. The fact that the strict epistemic pooling principle is satisfied for the scoring functions $\gamma_p$ thus implies that the weak epistemic pooling principle is satisfied for the modified scoring functions $\gamma'_p$, while we can still model every epistemic state, i.e.\ exhaustiveness \eqref{eqAvgAssumptionEveryCombinationWeak} is also satisfied. 

The discontinuous nature of the scoring function $\gamma'_p$ defined in \eqref{eqAvgWeakStrictDuality1} is clearly undesirable in practice. Hence the question arises whether it is possible to satisfy the weak epistemic pooling principle when only continuous scoring functions can be used. The answer to this question is negative. In particular, as the following result shows, if the weak epistemic pooling principle is satisfied with continuous scoring functions, all embeddings encode the same epistemic state.

\begin{proposition}\label{propAvgBoundaryWeak}
Suppose the epistemic pooling principle \eqref{eqEpistemicPoolingWeak} is satisfied for all embeddings $\mathbf{e},\mathbf{f}\in X$, for $\diamond=\diamond_{\mathsf{avg}}$ and $X$ convex. Suppose $\gamma_p$ is continuous. It holds that either $\mathsf{Neg}'_p =\emptyset$ or $\mathsf{Pos}'_p =\emptyset$.
\end{proposition}
\begin{proof}
Suppose there exists some $\mathbf{e}\in \mathsf{Neg}'_p$ and $\mathbf{f}\in \mathsf{Pos}'_p$. We then have $\gamma_p(\mathbf{e})< 0$  and $\gamma_p(\mathbf{f})\geq 0$. For $\lambda\in[0,1]$ we define:
$$
\mathbf{x}_{\lambda} = \lambda\mathbf{f} + (1-\lambda)\mathbf{e}
$$
By repeatedly applying \eqref{eqEpistemicPoolingWeak}, we find that $\gamma_p(\mathbf{x}_{\frac{1}{2^i}})\geq 0$ for every $i\in \mathbb{N}$. In particular, this means that for every $\varepsilon>0$ there exists some $\mathbf{x}\in X$ such that $d(\mathbf{e},\mathbf{x})<\varepsilon$ and $\gamma_p(\mathbf{x})\geq 0$. If $\gamma_p$ is continuous this implies $\gamma_p(\mathbf{e})=0$, which is a contradiction since we assumed $\mathbf{e}\in \mathsf{Neg}'_p$.
\end{proof}

\subsection{Summation}\label{secRealisabilitySummation}

\paragraph{Strict semantics}
We first show that the epistemic pooling principle \eqref{eqEpistemicPoolingGeneral} cannot be satisfied (in a non-trivial way) for all $\mathbf{e},\mathbf{f}\in \mathbb{R}^n$ when using $\diamond_{\mathsf{sum}}$, as we also found for $\diamond_{\mathsf{avg}}$ in Proposition \ref{propAvgWeakNeedsX}. 

\begin{proposition}
Suppose the epistemic pooling principle \eqref{eqEpistemicPoolingGeneral} is satisfied for all embeddings $\mathbf{e},\mathbf{f}\in \mathbb{R}^n$, with $\diamond=\diamond_{\mathsf{sum}}$. For any given $p\in \mathcal{P}$ we have
$$
(\forall \mathbf{e}\in\mathbb{R}^n\,.\, p\in \Gamma(\mathbf{e})) \vee (\forall \mathbf{e}\in\mathbb{R}^n\,.\, p\notin \Gamma(\mathbf{e}))
$$
\end{proposition}
\begin{proof}
Suppose $p\in \Gamma(\mathbf{e})$ and let $\mathbf{f}\in\mathbb{R}^n$. Using \eqref{eqEpistemicPoolingGeneral}, we find
$$
\Gamma(\mathbf{f}) = \Gamma(\mathbf{e} \diamond_{\mathsf{sum}} (\mathbf{f} - \mathbf{e})) = \Gamma(\mathbf{e})\cup \Gamma(\mathbf{f} - \mathbf{e})
$$
and thus $p\in \Gamma(\mathbf{f})$.
\end{proof}

We thus again need to define a suitable subset $X\subseteq \mathbb{R}^n$. To ensure that  $X$ is closed under $\diamond_{\mathsf{sum}}$ it is not sufficient that $X$ is convex. For this reason, we will assume that $X$ is conically closed, in particular:
\begin{align}\label{eqAssumptionConicallyClosed}
\forall \mathbf{e},\mathbf{f}\in X\,.\, \forall \alpha,\beta\geq 0\,.\, \alpha\,\mathbf{e} + \beta\,\mathbf{f} \in X
\end{align}
We now show that whenever \eqref{eqEpistemicPoolingGeneral} is satisfied for $\diamond_{\mathsf{sum}}$, for all $\mathbf{e},\mathbf{f}\in X$, it also holds that \eqref{eqEpistemicPoolingGeneral} is satisfied for $\diamond_{\mathsf{avg}}$, meaning that the results we have established for $\diamond_{\mathsf{avg}}$ carry over to $\diamond_{\mathsf{sum}}$. We first show the following lemma.

\begin{lemma}\label{lemmaSumScalingInvariant}
Suppose the epistemic pooling principle \eqref{eqEpistemicPoolingGeneral} is satisfied for all embeddings $\mathbf{e},\mathbf{f}\in X$, with $\diamond=\diamond_{\mathsf{sum}}$ and $X$ satisfying \eqref{eqAssumptionConicallyClosed}.
Let $p\in\mathcal{P}$ and $\mathbf{e}\in X$. If $\gamma_p(\mathbf{e})>0$ then it holds that $\gamma_p(\lambda\,\mathbf{e})>0$ for every $\lambda>0$. 
\end{lemma}
\begin{proof}
First note that $\gamma_p(\frac{\mathbf{e}}{2})> 0$. Indeed,  $\gamma_p(\frac{\mathbf{e}}{2})\leq 0$ would imply $\gamma_p(\mathbf{e})\leq 0$, given that \eqref{eqEpistemicPoolingGeneral} implies $\Gamma(\mathbf{e})=\Gamma(\frac{\mathbf{e}}{2})\cup \Gamma(\frac{\mathbf{e}}{2}) = \Gamma(\frac{\mathbf{e}}{2})$. Repeating the same argument,  we find $\gamma_p(\frac{\mathbf{e}}{2^i})> 0$ for any $i\in\mathbb{N}$. Hence, for every $\lambda>0$, there exists some $\lambda'\in]0,\lambda[$ such that $\gamma_p(\lambda'\mathbf{e})> 0$. Since $(\lambda-\lambda')\mathbf{e} \in X$, from \eqref{eqEpistemicPoolingGeneral} we then also find that $\gamma_p(\lambda\mathbf{e})>0$.
\end{proof}

\begin{proposition}\label{propSumPoolingThenAvgPooling}
Suppose the epistemic pooling principle \eqref{eqEpistemicPoolingGeneral} is satisfied for all embeddings $\mathbf{e},\mathbf{f}\in X$ with $\diamond=\diamond_{\mathsf{sum}}$ and $X$ satisfying \eqref{eqAssumptionConicallyClosed}. Then it also holds that \eqref{eqEpistemicPoolingGeneral} is satisfied for all embeddings $\mathbf{e},\mathbf{f}\in X$ with $\diamond=\diamond_{\mathsf{avg}}$.
\end{proposition}
\begin{proof}
Suppose \eqref{eqEpistemicPoolingGeneral} is satisfied for all embeddings $\mathbf{e},\mathbf{f}\in X$ with $\diamond=\diamond_{\mathsf{sum}}$. Let $\mathbf{e},\mathbf{f}\in X$ be such that $p\in \Gamma(\mathbf{e}\diamond_{\mathsf{avg}}\mathbf{f})$. In other words $\gamma_p(\frac{1}{2} (\mathbf{e}+\mathbf{f}))>0$. Using Lemma \ref{lemmaSumScalingInvariant} we then find  $\gamma_p(\mathbf{e}+\mathbf{f})>0$, and thus $p\in \Gamma(\mathbf{e}+\mathbf{f})$. Since \eqref{eqEpistemicPoolingGeneral} is satisfied for $\diamond_{\textit{sum}}$ we find $p\in \Gamma(\mathbf{e})\cup \Gamma(\mathbf{f})$.  
Conversely, assume that $p\in \Gamma(\mathbf{e})\cup \Gamma(\mathbf{f})$. Since \eqref{eqEpistemicPoolingGeneral} is satisfied for $\diamond_{\textit{sum}}$, this implies $p\in \Gamma(\mathbf{e}+\mathbf{f})$, and using Lemma \ref{lemmaSumScalingInvariant} we find $p\in \Gamma(\frac{1}{2}(\mathbf{e}+\mathbf{f}))$.
\end{proof}

Among others, it follows from Proposition \ref{propSumPoolingThenAvgPooling} that whenever \eqref{eqEpistemicPoolingGeneral} is satisfied for all embeddings $\mathbf{e},\mathbf{f}\in X$ with $\diamond=\diamond_{\mathsf{sum}}$, we have that $\mathsf{Pos}_p$ and $\mathsf{Neg}_p$ are convex for every $p\in \mathcal{P}$, and $\mathsf{Neg}_p\subseteq \delta(X)$. In particular, we also have the following corollary.

\begin{corollary}\label{corDimSum1}
Suppose the epistemic pooling principle \eqref{eqEpistemicPoolingGeneral} is satisfied for all embeddings $\mathbf{e},\mathbf{f}\in X$, with $\diamond=\diamond_{\mathsf{sum}}$ and $X\subseteq \mathbb{R}^n$ satisfying \eqref{eqAssumptionConicallyClosed}. Suppose that exhaustiveness \eqref{eqAvgAssumptionEveryCombination} is satisfied. It holds that $n\geq \vert\mathcal{P}\vert$.
\end{corollary}

\paragraph{Weak semantics}
In entirely the same way as Proposition \ref{propSumPoolingThenAvgPooling}, we can show the following result.

\begin{proposition}\label{propSumPoolingThenAvgPoolingWeak}
Suppose the epistemic pooling principle \eqref{eqEpistemicPoolingWeak} is satisfied for all embeddings $\mathbf{e},\mathbf{f}\in X$ with $\diamond=\diamond_{\mathsf{sum}}$ and $X$ satisfying \eqref{eqAssumptionConicallyClosed}. Then it also holds that \eqref{eqEpistemicPoolingWeak} is satisfied for all embeddings $\mathbf{e},\mathbf{f}\in X$ with $\diamond=\diamond_{\mathsf{avg}}$.
\end{proposition}

This means that we have the same negative result as we found for $\diamond_{\mathsf{avg}}$. In particular, from Propositions \ref{propAvgBoundaryWeak} and \ref{propSumPoolingThenAvgPoolingWeak}, we immediately obtain the following corollary.

\begin{corollary}
Suppose the epistemic pooling principle \eqref{eqEpistemicPoolingWeak} is satisfied for all embeddings $\mathbf{e},\mathbf{f}\in X$, for $\diamond=\diamond_{\mathsf{sum}}$ and $X$ satisfying \eqref{eqAssumptionConicallyClosed}. Suppose $\gamma_p$ is continuous. It holds that either $\mathsf{Neg}'_p =\emptyset$ or $\mathsf{Pos}'_p =\emptyset$.
\end{corollary}

\subsection{Max-pooling}\label{secPoolingPropertiesMax}

\begin{figure}
\centering
\includegraphics[width=175pt]{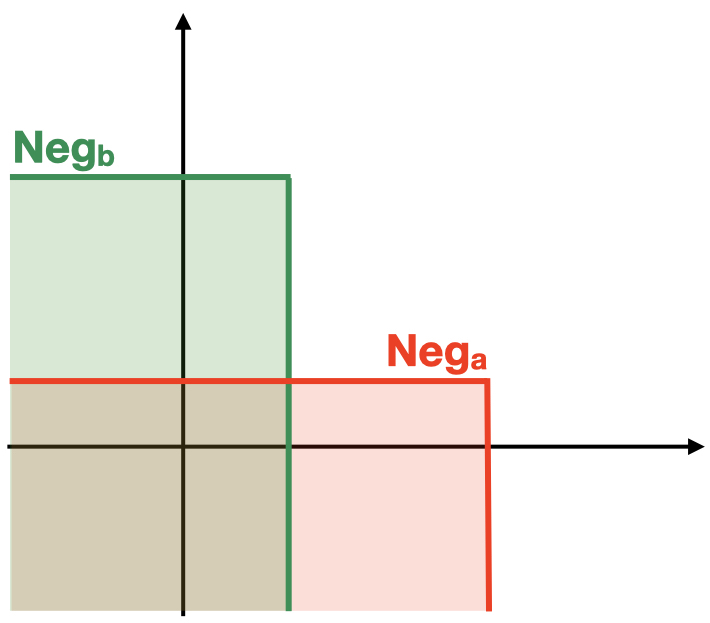}
\caption{Construction satisfying \eqref{eqEpistemicPoolingGeneral} and \eqref{eqAvgAssumptionEveryCombination} for $\diamond_{\mathsf{max}}$. \label{figMaxPoolingConstruction}}
\end{figure}

\paragraph{Strict semantics}
In contrast to what we found for $\diamond_{\mathsf{avg}}$ and $\diamond_{\mathsf{sum}}$, when using $\diamond_{\mathsf{max}}$ it is possible to satisfy the epistemic pooling principle \eqref{eqEpistemicPoolingGeneral} for all $\mathbf{e},\mathbf{f}\in \mathbb{R}^n$ in a non-trivial way. The main idea is illustrated in Figure \ref{figMaxPoolingConstruction}. For reasons that will become clear in Section \ref{secPropositionalReasoning}, in addition to the case where $X=\mathbb{R}^n$, we also consider the case where $X=]-\infty,z]^n$ for some $z\in\mathbb{R}$. We now first show the following characterisation: whenever \eqref{eqEpistemicPoolingGeneral} is satisfied in a non-trivial way, we always have that $\mathsf{Neg}_p$ is of the form $X \cap (Y_1\times...\times Y_n)$, where each $Y_i$ is of the form $]-\infty,b_i[$, $]-\infty,b_i]$ or $]-\infty,+\infty[$. Before we show this result, we show a number of lemmas.

\begin{lemma}\label{lemmaMaxPoolingInequalityA}
Let $X=\mathbb{R}^n$ or $X=]-\infty,z]^n$.
Suppose that the epistemic pooling principle \eqref{eqEpistemicPoolingGeneral} is satisfied for every $\mathbf{e},\mathbf{f}\in X$, with $\diamond=\diamond_{\mathsf{max}}$.
Let $\mathbf{x}=(x_1,...,x_n)\in X$ and $\mathbf{y}=(y_1,...,y_n)\in X$ be such that $\forall i\in\{1,...,n\}\,.\, x_i\leq y_i$. Then the following implication holds for every $p\in\mathcal{P}$:
\begin{align}
\mathbf{y}\in \mathsf{Neg}_p &\Rightarrow \mathbf{x}\in \mathsf{Neg}_p
\label{eqLemmaMaxPoolingImplication2}
\end{align}
\end{lemma}
\begin{proof}
Given that $\mathbf{x}\diamond_{\mathsf{max}}\mathbf{y}=\mathbf{y}$, it follows from \eqref{eqEpistemicPoolingGeneral} that $\Gamma(\mathbf{x})\subseteq \Gamma(\mathbf{y})$, from which we immediately find \eqref{eqLemmaMaxPoolingImplication2}.
\end{proof}

\begin{lemma}\label{lemmaMaxPoolingClosure2}
Let $X=\mathbb{R}^n$ or $X=]-\infty,z]^n$.
Suppose that the epistemic pooling principle \eqref{eqEpistemicPoolingGeneral} is satisfied for every $\mathbf{e},\mathbf{f}\in X$, with $\diamond=\diamond_{\mathsf{max}}$.
Let $\mathbf{x}=(x_1,...,x_n)\in X$ and $\mathbf{y}=(y_1,...,y_n)\in X$ be such that $\forall i\in\{1,...,n\}\,.\, x_i\leq y_i$. Then the following implication holds for every $p\in\mathcal{P}$:
\begin{align*}
\mathbf{y}\in \textit{cl}(\mathsf{Neg}_p) \Rightarrow \mathbf{x}\in \textit{cl}(\mathsf{Neg}_p)
\end{align*}
\end{lemma}
\begin{proof}
Suppose $\mathbf{y}\in \textit{cl}(\mathsf{Neg}_p)$. We need to show that for every $\varepsilon>0$ there exists some $\mathbf{x'}\in \mathsf{Neg}_p$ such that $d(\mathbf{x},\mathbf{x'})<\varepsilon$. Let $\varepsilon>0$. Since $\mathbf{y}\in \textit{cl}(\mathsf{Neg}_p)$, there exists some $\mathbf{y'}=(y'_1,...,y'_n)\in \mathsf{Neg}_p$ such that $d(\mathbf{y},\mathbf{y'})<\varepsilon$. Due to the assumption that $\forall i\in\{1,...,n\}\,.\, x_i\leq y_i$, we also have $\forall i\in\{1,...,n\}\,.\, y'_i - y_i + x_i\leq y'_i$. Using Lemma \ref{lemmaMaxPoolingInequalityA}, this implies $\mathbf{y'}-\mathbf{y}+\mathbf{x} \in \mathsf{Neg}_p$. Moreover, we have $d(\mathbf{x},\mathbf{y'}-\mathbf{y}+\mathbf{x})=d(\mathbf{y'},\mathbf{y})<\varepsilon$. It thus follows that $\mathbf{x}\in \textit{cl}(\mathsf{Neg}_p)$.
\end{proof}

\begin{lemma}\label{lemmaMaxPoolingClosure1}
Let $X=\mathbb{R}^n$ or $X=]-\infty,z]^n$.
Suppose that the epistemic pooling principle \eqref{eqEpistemicPoolingGeneral} is satisfied for every $\mathbf{e},\mathbf{f}\in X$, with $\diamond=\diamond_{\mathsf{max}}$.
If $\mathbf{x},\mathbf{y}\in \textit{cl}(\mathsf{Neg}_p)$ then we also have $\mathbf{x}\diamond_{\mathsf{max}}\mathbf{y}\in \textit{cl}(\mathsf{Neg}_p)$.
\end{lemma}
\begin{proof}
Let $\mathbf{x},\mathbf{y}\in \textit{cl}(\mathsf{Neg}_p)$.
We show that for every $\varepsilon>0$, there is some $\mathbf{e}\in \mathsf{Neg}_p$ such that $d(\mathbf{x}\diamond_{\mathsf{max}}\mathbf{y},\mathbf{e})<\varepsilon$. Since $\mathbf{x},\mathbf{y}\in \textit{cl}(\mathsf{Neg}_p)$, there exist $\mathbf{e_x},\mathbf{e_y}\in \mathsf{Neg}_p$ such that $d(\mathbf{e_x},\mathbf{x}) < \frac{\varepsilon}{\sqrt{n}}$ and $d(\mathbf{e_y},\mathbf{y}) < \frac{\varepsilon}{\sqrt{n}}$. Since $\mathbf{e_x},\mathbf{e_y}\in \mathsf{Neg}_p$, by  \eqref{eqEpistemicPoolingGeneral}  we also have $\mathbf{e_x}\diamond_{\mathsf{max}} \mathbf{e_y}\in \mathsf{Neg}_p$. Moreover, we have, for $\mathbf{e_x}=(e_{x,1},...,e_{x,n})$, $\mathbf{e_y}=(e_{y,1},...,e_{y,n})$, $\mathbf{x}=(x_{1},...,x_{n})$ and $\mathbf{y}=(y_{1},...,y_{n})$:
\begin{align*}
d^2(\mathbf{e_x}\diamond_{\mathsf{max}} \mathbf{e_y},\mathbf{x}\diamond_{\mathsf{max}}\mathbf{y})
&= \sum_{i=1}^n (\max(e_{x,i},e_{y,i})-\max(x_i,y_i))^2
< \sum_{i=1}^n \left(\frac{\varepsilon}{\sqrt{n}}\right)^2
= \varepsilon^2
\end{align*}
and thus $d(\mathbf{e_x}\diamond_{\mathsf{max}} \mathbf{e_y},\mathbf{x}\diamond_{\mathsf{max}}\mathbf{y}) < \varepsilon$.
\end{proof}

\begin{proposition}\label{propMaxPoolingBoxes}
Let $X=\mathbb{R}^n$ or $X=]-\infty,z]^n$.
Suppose that the epistemic pooling principle \eqref{eqEpistemicPoolingGeneral} is satisfied for every $\mathbf{e},\mathbf{f}\in X$, with $\diamond=\diamond_{\mathsf{max}}$. Let $p\in\mathcal{P}$.
It holds that $\textit{cl}(\mathsf{Neg}_p) = X \cap (Y_p^1\times ... \times Y_p^n)$ where for every $i\in\{1,...,n\}$, we have $Y_p^i=]-\infty,b_i]$ for some $b_i\in \mathbb{R}$ or $Y_p^i=]-\infty,+\infty[$. 
\end{proposition}
\begin{proof}
For each $i\in\{1,...,n\}$, we can consider two cases:
\begin{itemize}
\item Assume that the $i^{\text{th}}$ coordinate of the elements from $\textit{cl}(\mathsf{Neg}_p)$ is bounded, i.e.\ there exists some $b_i\in \mathbb{R}$ such that for each $(x_1,...,x_n)\in \textit{cl}(\mathsf{Neg}_p)$ it holds that $x_i\leq b_i$. Suppose $(x_1,...,x_n)$ and $(y_1,...,y_n)$ are elements from $\textit{cl}(\mathsf{Neg}_p)$ which are maximal in the $i^{\textit{th}}$ coordinate, i.e.\ for any $\varepsilon >0$ we have $(x_1,\allowbreak ...,\allowbreak x_{i-1},\allowbreak x_i+\varepsilon,\allowbreak x_{i+1},\allowbreak ...,\allowbreak x_n)\notin \textit{cl}(\mathsf{Neg}_p)$ and $(y_1,\allowbreak ...,\allowbreak y_{i-1},\allowbreak y_i+\varepsilon,\allowbreak y_{i+1},\allowbreak ...,\allowbreak y_n)\notin \textit{cl}(\mathsf{Neg}_p)$. Assume furthermore that $x_i<y_i$. 
We have that $(\max(x_1,y_1),\allowbreak ...,\allowbreak \max(x_n,y_n))\in \textit{cl}(\mathsf{Neg}_p)$ by Lemma \ref{lemmaMaxPoolingClosure1}, which implies $(x_1,...,x_{i-1},y,x_{i+1},...x_n)\in \textit{cl}(\mathsf{Neg}_p)$ by Lemma \ref{lemmaMaxPoolingClosure2}. However, this is in contradiction with the assumption we made about the $i^{\textit{th}}$ coordinate of $(x_1,...,x_n)$.
It follows that there is a constant $b_i\in \mathbb{R}$ such that for every $(x_1,...,x_n)\in  \textit{cl}(\mathsf{Neg}_p)$, we have 
$$
b_i=\max \{ y \,|\, (x_1,\allowbreak ...,\allowbreak x_{i-1},\allowbreak y,\allowbreak x_{i+1},\allowbreak ...,\allowbreak x_n) \in \textit{cl}(\mathsf{Neg}_p)\}
$$Moreover, from Lemma \ref{lemmaMaxPoolingClosure2} we  find that $(x_1,\allowbreak ...,\allowbreak x_{i-1},\allowbreak y,\allowbreak x_{i+1},\allowbreak ...,\allowbreak x_n)\in \textit{cl}(\mathsf{Neg}_p)$ for every $y\in ]-\infty,b_i]$.
\item Now we consider the case where the $i^{\text{th}}$ coordinate of the elements from $\textit{cl}(\mathsf{Neg}_p)$ is unbounded. Let $(y_1,...,y_n)\in \textit{cl}(\mathsf{Neg}_p)$. We show that for any $z\in\mathbb{R}$ it holds that $(y_1,...,y_{i-1},z,y_{i+1},...,y_n)\in \textit{cl}(\mathsf{Neg}_p)$. Since we assumed the $i^{\text{th}}$ coordinate of the elements from $\textit{cl}(\mathsf{Neg}_p)$ is unbounded, there exists some $(x_1,...,x_n)\in \textit{cl}(\mathsf{Neg}_p)$ such that $x_i\geq z$. 
From $(x_1,\allowbreak,...,x_n)\in \textit{cl}(\mathsf{Neg}_p)$  and $(y_1, ...,\allowbreak y_n)\in\textit{cl}(\mathsf{Neg}_p)$ we find $(\max(x_1,y_1),..., \max(x_n,y_n))\in \textit{cl}(\mathsf{Neg}_p)$, using Lemma \ref{lemmaMaxPoolingClosure1}, which in turn implies $(y_1,\allowbreak ...,\allowbreak y_{i-1},\allowbreak z,\allowbreak y_{i+1},\allowbreak ...,\allowbreak y_n)\in \textit{cl}(\mathsf{Neg}_p)$, using Lemma \ref{lemmaMaxPoolingClosure2}.
\end{itemize}
Putting these two cases together, find that $\textit{cl}(\mathsf{Neg}_p)$ is of the form $X\cap (Y_p^1\times ... Y_p^n)$.  
\end{proof}

Using the characterisation from Proposition \ref{propMaxPoolingBoxes}, we now show that embeddings with a minimum of $|\mathcal{P}|$ dimensions are needed to satisfy the epistemic pooling principle.

\begin{proposition}\label{eqLowerboundMaxPoolingStrict}
Let $X=\mathbb{R}^n$ or $X=]-\infty,z]^n$.
Suppose the epistemic pooling principle \eqref{eqEpistemicPoolingGeneral} is satisfied for all embeddings $\mathbf{e},\mathbf{f}\in X$, with $\diamond=\diamond_{\mathsf{max}}$. Suppose that exhaustiveness \eqref{eqAvgAssumptionEveryCombination} is satisfied. It holds that $n\geq \vert\mathcal{P}\vert$.
\end{proposition}
\begin{proof}
From Proposition \ref{propMaxPoolingBoxes} we know that for each property $p\in \mathcal{P}$, it holds that $\textit{cl}(\mathsf{Neg}_p)$ is of the form $X \cap (Y_p^1\times ... \times Y_p^n)$, where $Y_p^i=]-\infty,b_i]$ or $Y_p^i=]-\infty,+\infty[$. Let us write $p <_i q$ for $p,q\in\mathcal{P}$ to denote that one of the following cases holds:
\begin{itemize}
\item  $Y_p^i=]-\infty,b_i^p]$ and $Y_q^i=]-\infty,b_i^q]$ with $b_i^p < b_i^q$; or
\item $Y_p^i=]-\infty,b_i^p]$ and $Y_q^i=]-\infty,b_i^q]$ with $b_i^p = b_i^q=b$ and there exists an element $(x_1,...,x_n)\in \mathsf{Neg}_q$ such that $x_i=b$ while no such element exists in $\mathsf{Neg}_p$. In other words, the upper bound $b$ for the $i^{\textit{th}}$ coordinate is strict for $\mathsf{Neg}_p$ but not for $\mathsf{Neg}_q$.
\item  $Y_p^i=]-\infty,b_i^p]$ and $Y_q^i= ]-\infty,+\infty[$.
\end{itemize}
For each $i\in\{1,....,n\}$, we can choose a property $p_i$ from $\mathcal{P}$ which is minimal w.r.t.\ the relation $<_i$. Suppose there was some property $q\in \mathcal{P}\setminus\{p_1,...,p_n\}$. Let $\mathbf{x}=(x_1,...,x_n)\in \mathsf{Pos}_q$. Then for some coordinate $i$, it must be the case that $Y_q^i=]-\infty,b_i^q]$ and either (i) $x_i>b_i^q$ or (ii) $x_i=b_i^q$ but $b_i^q$ represents a strict upper bound for the $i^{\textit{th}}$ coordinate. 
Because $p_i$ was chosen as a minimal element w.r.t.\ $<_i$ it follows that $(x_1,...,x_n)\notin \mathsf{Neg}_{p_i}$. We thus find that for every $\mathbf{x}\in \mathsf{Pos}_q$ it holds that $\mathbf{x}\in \mathsf{Pos}_{p_1}\cup...\cup \mathsf{Pos}_{p_n}$. It follows that there is no $\mathbf{x}\in\mathbb{R}^n$ such that $\Gamma(\mathbf{x})=\{q\}$, meaning that  \eqref{eqAvgAssumptionEveryCombination} is not satisfied.
\end{proof}

To show that the epistemic pooling principle \eqref{eqEpistemicPoolingGeneral} and exhaustiveness \eqref{eqAvgAssumptionEveryCombination} can indeed be satisfied with $\vert\mathcal{P}\vert$ dimensions, let $\mathcal{P}=\{p_1,...,p_n\}$ and define:
\begin{align}\label{eqScoringStrictMaxPooling}
\gamma_{p_i}(e_1,...,e_n)=e_i
\end{align}
Then we have that the $i\textsuperscript{th}$ coordinate of $(e_1,...,e_n)\diamond_{\mathsf{max}}(f_1,...,f_n)$ is strictly positive iff $e_i>0$ or $f_i>0$, hence we indeed have $\Gamma(\mathbf{e}\diamond_{\mathsf{max}}\mathbf{f})=\Gamma(\mathbf{e})\cup \Gamma(\mathbf{f})$, meaning that the epistemic pooling principle \eqref{eqEpistemicPoolingGeneral} is satisfied for every $\mathbf{e},\mathbf{f}\in \mathbb{R}^n$. To see why exhaustiveness \eqref{eqAvgAssumptionEveryCombination} is satisfied, let $\mathcal{Q}\subseteq \mathcal{P}$. We define $\mathbf{q}=(q_1,...,q_n)$ as follows:
\begin{align*}
q_i = 
\begin{cases}
1 & \text{if $p_i\in\mathcal{Q}$}\\
-1 & \text{otherwise}
\end{cases}
\end{align*}

\paragraph{Weak semantics}
As before, the main question is whether it is possible to satisfy the weak epistemic pooling principle \eqref{eqEpistemicPoolingWeak} in a non-trivial way using continuous scoring functions $\gamma_p$, since the results from the strict semantics trivially carry over to the weak semantics if non-continuous scoring functions are allowed. This is indeed the case. In fact, with the scoring functions defined in \eqref{eqScoringStrictMaxPooling}, the weak epistemic pooling principle is also satisfied. Moreover, in the same way as for the strict semantics, we find that exhaustiveness \eqref{eqAvgAssumptionEveryCombination} is satisfied for this choice. Finally, note that the lower bound $n\geq \vert\mathcal{P}\vert$ still applies for the weak semantics, which can be shown in exactly the same way as Proposition \ref{eqLowerboundMaxPoolingStrict}.

\subsection{Hadamard product}\label{secPoolingPropertiesHad}

\begin{figure}
\centering
\includegraphics[width=175pt]{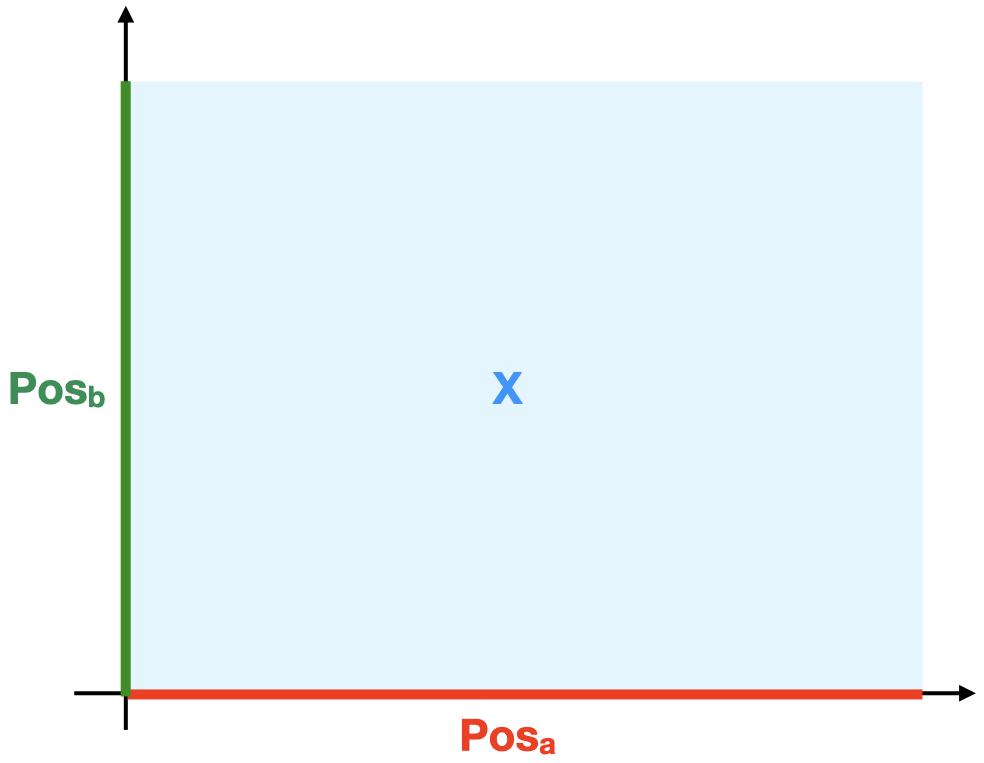}
\caption{Construction satisfying \eqref{eqEpistemicPoolingGeneral} and \eqref{eqAvgAssumptionEveryCombination} for $\diamond_{\mathsf{had}}$ with $X=[0,+\infty[^n$. \label{figHadPoolingConstruction}}
\end{figure}

\paragraph{Strict semantics}
Similar as we found for max-pooling, with the Hadamard product $\diamond_{\mathsf{had}}$, it is possible to satisfy the epistemic pooling principle \eqref{eqEpistemicPoolingGeneral} for every $\mathbf{e},\mathbf{f}\in \mathbb{R}^n$, while also satisfying exhaustiveness \eqref{eqAvgAssumptionEveryCombination}. In addition to the choice $X=\mathbb{R}^n$, we also consider the case where $X=[0,+\infty[^n$. As we will see, the results we establish in this section are valid regardless of whether $X=\mathbb{R}^n$ or $X=[0,+\infty[^n$. The reason why we specifically include the case $X=[0,+\infty[^n$ will become clear in Section \ref{secPropositionalReasoning}. Figure \ref{figHadPoolingConstruction} illustrates how the epistemic pooling principle \eqref{eqEpistemicPoolingGeneral} can be satisfied for the case where $X=[0,+\infty[^n$. Note the dual nature of this construction and the construction for $\diamond_{\mathsf{avg}}$ from Figure \ref{figAvgPoolingConstruction}, where the positive regions in Figure \ref{figHadPoolingConstruction} correspond to the negative regions in Figure \ref{figAvgPoolingConstruction}.
Let us write $H_i$ for the following hyperplane:
$$
H_i = \{(x_1,...,x_n) \in \mathbb{R}^n \mid x_i=0\}
$$
These hyperplanes play a particular role in the characterisation of the positive regions $\mathsf{Pos}_p$, as was already illustrated in Figure \ref{figHadPoolingConstruction}. The following results make this role explicit.

\begin{lemma}\label{lemmaHadIndexSet1}
Let $X=\mathbb{R}^n$ or $X=[0,+\infty[^n$. Suppose the epistemic pooling principle \eqref{eqEpistemicPoolingGeneral} is satisfied for all embeddings $\mathbf{e},\mathbf{f}\in X$, with $\diamond=\diamond_{\mathsf{had}}$. Let $p\in \mathcal{P}$ and $\mathbf{e}=(e_1,...,e_n) \in\mathsf{Pos}_p$. Let $I=\{i\in \{1,...,n\} \mid e_i=0\}$. It holds that
$$
X \cap \bigcap_{i\in I} H_i \subseteq \mathsf{Pos}_p
$$
\end{lemma}
\begin{proof}
Let $\mathbf{f}=(f_1,...,f_n)\in X \cap \bigcap_{i\in I} H_i$. We show that $\mathbf{f}\in \mathsf{Pos}_p$.  Let $\mathbf{x}=(x_1,...,x_n)$ be defined as follows:
\begin{align*}
x_i =
\begin{cases}
\frac{f_i}{e_i} & \text{if $e_i\neq 0$}\\
0               & \text{otherwise}
\end{cases}
\end{align*}
Since $(f_1,...,f_n)\in \bigcap_{i\in I} H_i$ it holds that $f_i=0$ whenever $e_i=0$. We thus have $\mathbf{f}=\mathbf{e}\diamond_{\mathsf{had}}\mathbf{x}$. Using \eqref{eqEpistemicPoolingGeneral}, it then follows from $\mathbf{e}\in \mathsf{Pos}_p$ that $\mathbf{f}\in \mathsf{Pos}_p$.
\end{proof}

It follows that $\mathsf{Pos}_p$ is a finite union of regions of the form $X \cap \bigcap_{i\in I} H_i$. In particular, it also follows that $\textit{dim}(\mathsf{Pos}_p)\leq n-1$ if the epistemic pooling principle \eqref{eqEpistemicPoolingGeneral} is satisfied in a non-trivial way.
For a given index set $I\subseteq \{1,...,n\}$, let us define:
$$
\mathcal{P}_I = \{p\in \mathcal{P} \mid X \cap \bigcap_{i\in I} H_i \subseteq \mathsf{Pos}_p\}
$$
\begin{lemma}\label{lemmaHadIndexSet2}
Let $X=\mathbb{R}^n$ or $X=[0,+\infty[^n$. Suppose the epistemic pooling principle \eqref{eqEpistemicPoolingGeneral} is satisfied for all embeddings $\mathbf{e},\mathbf{f}\in X$, with $\diamond=\diamond_{\mathsf{had}}$. 
Let $I,J \subseteq \{1,...,n\}$. It holds that: 
$$
\mathcal{P}_{I\cup J} = \mathcal{P}_I \cup \mathcal{P}_J
$$
\end{lemma}
\begin{proof}
We clearly have $\mathcal{P}_I \subseteq \mathcal{P}_{I\cup J}$, since $X \cap \bigcap_{i\in I} H_i \subseteq \mathsf{Pos}_p$ implies $X \cap \bigcap_{i\in I\cup J} H_i \subseteq \mathsf{Pos}_p$. Similarly, we have $\mathcal{P}_J \subseteq \mathcal{P}_{I\cup J}$ and thus we find $\mathcal{P}_I \cup \mathcal{P}_J\subseteq \mathcal{P}_{I\cup J}$. We now show the other direction. Suppose $p\in \mathcal{P}_{I\cup J}$. Let $\mathbf{e}=(e_1,...,e_n)$ where $e_i=0$ if $i\in I\cup J$ and $e_i=1$ otherwise. Since $p\in \mathcal{P}_{I\cup J}$ we have $\mathbf{e}\in \mathsf{Pos}_p$. Now let $\mathbf{f}=(f_1,...,f_n)$ where $f_i=1$ if $i\in I$ and $f_i=0$ otherwise. Similarly, let $\mathbf{g}=(g_1,...,g_n)$ where $g_i=1$ if $i\in J$ and $g_i=0$ otherwise. Then $\mathbf{e}=\mathbf{f}\odot\mathbf{g}$. From \eqref{eqEpistemicPoolingGeneral} it follows that $\mathbf{f}\in \mathsf{Pos}_p$ or $\mathbf{g}\in \mathsf{Pos}_p$. Thus, using Lemma \ref{lemmaHadIndexSet1} we find that $p\in \mathcal{P}_I$ or $p\in \mathcal{P}_J$.
\end{proof}

We can now show that at least $|\mathcal{P}|$ dimensions are again needed to satisfy \eqref{eqEpistemicPoolingGeneral} in a nontrivial way. 

\begin{proposition}\label{propMinDimHadStrict}
Let $X=\mathbb{R}^n$ or $X=[0,+\infty[^n$. Suppose the epistemic pooling principle \eqref{eqEpistemicPoolingGeneral} is satisfied for all embeddings $\mathbf{e},\mathbf{f}\in X$, with $\diamond=\diamond_{\mathsf{had}}$. Suppose that exhaustiveness \eqref{eqAvgAssumptionEveryCombination} is satisfied. It holds that $n\geq \vert\mathcal{P}\vert$.
\end{proposition}
\begin{proof}
Given \eqref{eqAvgAssumptionEveryCombination}, for each $p\in \mathcal{P}$, there must exist some $(e^p_1,...,e^p_n)\in X$ such that $\Gamma(e^p_1,...,e^p_n)=\{p\}$. Let us fix such vectors $(e^p_1,...,e^p_n)$ for each $p\in \mathcal{P}$ and define $I_p = \{i\in\{1,...,n\} \mid e^p_i=0\}$. Note that by Lemma \ref{lemmaHadIndexSet1}, we have $X\cap \bigcap_{i\in I_p}H_i \subseteq \mathsf{Pos}_{p}$. Moreover, since $\Gamma(e^p_1,...,e^p_n)=\{p\}$, we have $X\cap \bigcap_{i\in I_p}H_i \not\subseteq \mathsf{Pos}_{q}$ for any $q\neq p$. In other words, we have $\mathcal{P}_{I_p}=\{p\}$. 

For $p\neq q$ we clearly have $I_p\not\subseteq I_q$, since $I_p\subseteq I_q$ would imply $\mathcal{P}_{I_q}=\{p,q\}$. This implies in particular that $I_p\neq \emptyset$ for every $p\in \mathcal{P}$. Now let us consider $k$ distinct properties $p_1,p_2,...,p_k$. Then we cannot have $I_{p_k}\subseteq I_{p_1}\cup I_{p_2}\cup ... \cup I_{p_{k-1}}$. Indeed $I_{p_k}\subseteq I_{p_1}\cup ...\cup I_{p_{k-1}}$ would imply $p_k\in \mathcal{P}_{I_{p_1}\cup ...\cup I_{p_{k-1}}}$ whereas from Lemma \ref{lemmaHadIndexSet2} we know that $\mathcal{P}_{I_{p_1}\cup ...\cup I_{p_{k-1}}}= \mathcal{P}_{I_{p_1}}\cup ... \cup \mathcal{P}_{I_{p_{k-1}}}$ and we know that the latter is equal to $\{p_1,...,p_{k-1}\}$. In other words, there is at least one element in $I_{p_k}$ which does not occur in $I_{p_1},...,I_{p_{k-1}}$, a contradiction. Since this needs to hold for every $k\in\{2,...,\vert\mathcal{P}\vert\}$, there need to be at least $\vert\mathcal{P}\vert$ distinct elements in $I_{p_1}\cup ...\cup I_{p_{\vert\mathcal{P}\vert}}$. This means that $n\geq \vert\mathcal{P}\vert$.
\end{proof}

To show that the epistemic pooling principle \eqref{eqEpistemicPoolingGeneral} and exhaustiveness \eqref{eqAvgAssumptionEveryCombination} can indeed be satisfied with $\vert\mathcal{P}\vert$ dimensions, let $\mathcal{P}=\{p_1,...,p_n\}$ and define:
\begin{align}\label{eqScoringStrictHadPooling}
\gamma_{p_i}(e_1,...,e_n)=
\begin{cases}
1 & \text{if $e_i=0$}\\
0 & \text{otherwise}
\end{cases}
\end{align}
Clearly, the $i\textsuperscript{th}$ coordinate of $(e_1,...,e_n)\diamond_{\mathsf{had}}(f_1,...,f_n)$ is 0 iff $e_i=0$ or $f_i=0$, hence we indeed have $\Gamma(\mathbf{e}\diamond_{\mathsf{had}}\mathbf{f})=\Gamma(\mathbf{e})\cup \Gamma(\mathbf{f})$, meaning that \eqref{eqEpistemicPoolingGeneral} is satisfied for every $\mathbf{e},\mathbf{f}\in \mathbb{R}^n$. It is also straightforward to verify that exhaustiveness \eqref{eqAvgAssumptionEveryCombination} is satisfied. Note, however, that the scoring function $\gamma_{p_i}$ defined in \eqref{eqScoringStrictHadPooling} is not continuous. As the following result shows, this is not a coincidence.

\begin{proposition}\label{propHadStricContinuous}
Let $X=\mathbb{R}^n$ or $X=[0,+\infty[^n$.
Suppose the epistemic pooling principle \eqref{eqEpistemicPoolingGeneral} is satisfied for all embeddings $\mathbf{e},\mathbf{f}\in X$, for $\diamond=\diamond_{\mathsf{had}}$. Suppose $\gamma_p$ is continuous. It holds that either $\mathsf{Neg}_p =\emptyset$ or $\mathsf{Pos}_p =\emptyset$.
\end{proposition}
\begin{proof}
Suppose $\mathsf{Pos}_p \neq\emptyset$. Then from Lemma \ref{lemmaHadIndexSet1}, we know that there is some $I\subseteq \{1,...,n\}$ such that $X\cap \bigcap_{i\in I} H_i \subseteq \mathsf{Pos}_p$. Let us assume that $I$ is minimal, i.e.\ for any $I'\subset I$ we have $X\cap \bigcap_{i\in I'} H_i \not\subseteq \mathsf{Pos}_p$. Now suppose we also have $\mathsf{Neg}_p \neq\emptyset$. Then we have $I\neq \emptyset$. Let $i\in I$ and $\mathbf{e}_0 \in X\cap \bigcap_{i\in I} H_i$. For $\varepsilon>0$, let $\mathbf{e}_{\varepsilon}$ be the vector obtained from $\mathbf{e}_0$ by replacing the $i\textsuperscript{th}$ coordinate by $\varepsilon$. Since we assumed $I$ was minimal, for any $\varepsilon>0$ we have that $\mathbf{e}_{\varepsilon}\in \mathsf{Neg}_p$, or equivalently $\gamma_p(\mathbf{e}_{\varepsilon})\leq 0$. However, if $\gamma_p$ is continuous, this would imply $\gamma_p(\mathbf{e}_{0})\leq 0$ and thus $\mathbf{e}_{0}\in\mathsf{Neg}_p$, a contradiction.
\end{proof}

The fact that only discontinuous scoring functions can be used is an important limitation in practice. One solution is to make a different choice for the set $X$. For instance, if we choose $X=[1,+\infty[$, we end up with embeddings where the bounding hyperplanes, of the form $H=\{(e_1,...,e_n)\in\mathbb{R}^n \mid e_i=1\}$, correspond to negative regions. The resulting embeddings thus have similar characteristics as the ones we obtained with $\diamond_{\mathsf{avg}}$. For this reason, we will not consider this option in further detail.

\paragraph{Weak semantics}
Under the weak semantics, it is possible to satisfy the epistemic pooling principle using continuous scoring functions. For instance, if $\mathcal{P}=\{p_1,...,p_n\}$ we can define:
\begin{align}\label{eqScoringStrictHadPoolingWeak}
\gamma_{p_i}(e_1,...,e_n)= -e_i^2
\end{align}
It is straightforward to verify that, with this choice of scoring function, the epistemic pooling principle \eqref{eqEpistemicPoolingWeak} is satisfied for all $\mathbf{e},\mathbf{f}\in\mathbb{R}^n$, while exhaustiveness \eqref{eqAvgAssumptionEveryCombinationWeak} is also satisfied. Moreover, the same argument as in the proof of Proposition \ref{propMinDimHadStrict} can be used for the weak semantics as well, meaning that we still need at least $\vert\mathcal{P}\vert$ dimensions to satisfy \eqref{eqEpistemicPoolingWeak} with $\diamond=\diamond_{\mathsf{had}}$ in a non-trivial way.

\section{Propositional reasoning with epistemic embeddings}\label{secPropositionalReasoning}
Throughout this paper, we model epistemic states as subsets of $\mathcal{P}$. In general, we can simply think of the elements of $\mathcal{P}$ as atomic pieces of evidence. Crucially, however, this setting is expressive enough to capture propositional reasoning. In particular, suppose each embedding $\mathbf{e}$ is associated with a set $\Lambda(\mathbf{e})$ of propositional formulas. Now suppose the embedding $\mathbf{g}$ is obtained by pooling the embeddings $\mathbf{e}$ and $\mathbf{f}$, e.g.\ representing the information we have obtained from two different sources:
$$
\mathbf{g} = \mathbf{e}\diamond \mathbf{f}
$$
Then we want $\mathbf{g}$ to combine the knowledge captured by $\mathbf{e}$ and $\mathbf{f}$. In other words, we want $\Lambda(\mathbf{g})$ to be logically equivalent to $\Lambda(\mathbf{e})\cup\Lambda(\mathbf{f})$. This gives rise to the following variant of the epistemic pooling principle:
\begin{align}\label{eqPoolingPrincipleB}
\Lambda(\mathbf{e}\diamond\mathbf{f}) \equiv \Lambda(\mathbf{e})\cup \Lambda(\mathbf{e})
\end{align}
where we write $\equiv$ to denote logical equivalence between sets of propositional formulas. In other words, in this setting, we want to be able to do propositional reasoning by pooling embeddings. This can be achieved using our considered setting as follows. Let $\mathcal{W}$ be the set of all possible worlds, i.e.\ the set of all propositional interpretations over some set of atoms $\textit{At}$. We associate one property with each possible world:
\begin{align}\label{defPlogical}
\mathcal{P} = \{p_{\omega} \mid \omega\in\mathcal{W}\}
\end{align}
Intuitively, $p_{\omega}$ means that $\omega$ can be excluded, i.e.\ that we have evidence that $\omega$ is not the true world. Let $\textit{Clauses}$ be the set of all clauses over the considered set of propositional atoms $\textit{At}$. We can define $\Lambda(\mathbf{e})$ in terms of the scoring functions $\gamma_p$ as follows:
\begin{align}\label{defLambdalogical}
\Lambda(\mathbf{e}) = \{\alpha \in \textit{Clauses} \mid \forall \omega\in \mathcal{W}\,.\, (\gamma_{p_{\omega}}(\mathbf{e})\leq 0) \Rightarrow (\omega\models\alpha)\}
\end{align}
In other words, we have that $\alpha\in \Lambda(\mathbf{e})$ if $\alpha$ is true in all the worlds $\omega$ that cannot be excluded based on the evidence encoded by $\mathbf{e}$. Note that $\Lambda(\mathbf{e})$ is deductively closed, in the sense that $\alpha \in \Lambda(\mathbf{e})$ iff $\Lambda(\mathbf{e})\models \alpha$. We can also straightforwardly show the following characterisation:
\begin{lemma}\label{lemmaLambdaGammaConnection1}
Suppose $\mathcal{P}$ is defined by \eqref{defPlogical} and $\Lambda$ is defined by \eqref{defLambdalogical}. Let $p_{\omega}\in \mathcal{P}$. It holds that $\gamma_{p_{\omega}}(\mathbf{e})\leq 0$ iff $\omega\models \Lambda(\mathbf{e})$.
\end{lemma}
\begin{proof}
Suppose $\gamma_{p_{\omega}}(\mathbf{e})\leq 0$, then by definition of $\Lambda$, for each $\alpha\in\Lambda(\mathbf{e})$ we must have $\omega\models \alpha$. In particular, we must thus have $\omega\models \Lambda(\mathbf{e})$. 
Conversely, suppose $\gamma_{p_{\omega}}(\mathbf{e})> 0$. Let $\{\alpha_1,...,\alpha_k\}$ be a set of clauses whose set of models is given by $\{\omega' \,|\, \gamma_{p_{\omega'}}(\mathbf{e})\leq 0\}$. Note that such a set of clauses always exists. Moreover, we then have $\{\alpha_1,...,\alpha_k\}\subseteq \Lambda(\mathbf{e})$, by definition of $\Lambda$, while by construction we have $\omega\not \models \{\alpha_1,...,\alpha_n\}$. It follows that $\omega\not\models \Lambda(\mathbf{e})$.
\end{proof}

We can now prove the following result, which shows that \eqref{eqPoolingPrincipleB} arises as a special case of the (strict) epistemic pooling principle \eqref{eqEpistemicPoolingGeneral}.
\begin{proposition}\label{propFormulaPoolingEquivalence}
Suppose the epistemic pooling principle \eqref{eqEpistemicPoolingGeneral} is satisfied for all embeddings $\mathbf{e},\mathbf{f}\in X$, for some $X\subseteq \mathbb{R}^n$ and some pooling operator $\diamond$ such that $X$ is closed under $\diamond$. Suppose $\mathcal{P}$ is defined by \eqref{defPlogical} and $\Lambda$ is defined by \eqref{defLambdalogical}. It holds that \eqref{eqPoolingPrincipleB} is satisfied for all $\mathbf{e},\mathbf{f}\in X$.
\end{proposition}
\begin{proof}
Let $\mathbf{e},\mathbf{f}\in X$. By \eqref{eqEpistemicPoolingGeneral} we have that $\Gamma(\mathbf{e}\diamond\mathbf{f}) \supseteq \mathbf{e}$, hence for every $\omega$ we have:
$$(\gamma_{p_{\omega}}(\mathbf{e}\diamond\mathbf{f}) \leq 0) \Rightarrow (\gamma_{p_{\omega}}(\mathbf{e}) \leq 0) 
$$
From the definition of $\Lambda$, it follows that $\Lambda(\mathbf{e}\diamond\mathbf{f}) \supseteq \Lambda(\mathbf{e})$. Since we similarly have $\Lambda(\mathbf{e}\diamond\mathbf{f}) \supseteq \Lambda(\mathbf{f})$, we find $\Lambda(\mathbf{e}\diamond\mathbf{f}) \supseteq \Lambda(\mathbf{e})\cup  \Lambda(\mathbf{f})$
and in particular:
$$
\Lambda(\mathbf{e}\diamond\mathbf{f})\models \Lambda(\mathbf{e})\cup \Lambda(\mathbf{f})
$$
Conversely, suppose $\alpha \in \Lambda(\mathbf{e}\diamond\mathbf{f})$. Then we have
$
\forall \omega\in \mathcal{W}\,.\, (\gamma_{p_{\omega}}(\mathbf{e}\diamond\mathbf{f})\leq 0) \Rightarrow (\omega\models\alpha)
$.
From \eqref{eqEpistemicPoolingGeneral} we know that $\gamma_{p_{\omega}}(\mathbf{e}\diamond\mathbf{f})\leq 0$ iff $\gamma_{p_{\omega}}(\mathbf{e})\leq 0$ and $\gamma_{p_{\omega}}(\mathbf{f})\leq 0$, hence we obtain:
$$
\forall \omega\in \mathcal{W}\,.\, (\gamma_{p_{\omega}}(\mathbf{e})\leq 0) \wedge (\gamma_{p_{\omega}}(\mathbf{f})\leq 0) \Rightarrow (\omega\models\alpha)
$$
Using Lemma \ref{lemmaLambdaGammaConnection1} we find:
$$
\forall \omega\in \mathcal{W}\,.\, (\omega\models \Lambda(\mathbf{e})) \wedge (\omega\models \Lambda(\mathbf{f})) \Rightarrow (\omega\models\alpha)
$$
or, equivalently, $\forall \omega\in \mathcal{W}\,.\, (\omega\models \Lambda(\mathbf{e})\cup \Lambda(\mathbf{f})) \Rightarrow (\omega\models\alpha)$.
We thus find $\Lambda(\mathbf{e})\cup \Lambda(\mathbf{f}) \models \alpha$.
Since this holds for every $\alpha\in\Lambda(\mathbf{e}\diamond\mathbf{f})$, we find
$$
\Lambda(\mathbf{e})\cup \Lambda(\mathbf{f}) \models \Lambda(\mathbf{e}\diamond\mathbf{f})
$$
\end{proof}
We can similarly model propositional reasoning using the weak semantics, by defining the set of formulas associated with an embedding $\mathbf{e}$ as follows
\begin{align}\label{defLambdalogicalWeak}
\Lambda'(\mathbf{e}) = \{\alpha \in \textit{Clauses} \mid \forall \omega\in \mathcal{W}\,.\, (\gamma_{p_{\omega}}(\mathbf{e})< 0) \Rightarrow (\omega\models\alpha)\}
\end{align}
The counterpart to Proposition \ref{propFormulaPoolingEquivalence} for the weak semantics is shown in entirely the same way.
We can thus use the framework that was studied in Section \ref{secPoolingProperties} to combine, and reason about propositional knowledge. However, when we focus on knowledge that is encoded using propositional formulas, we also need an effective way to check whether a given formula $\alpha$ is entailed by the knowledge base $\Lambda(\mathbf{e})$, i.e.\ whether the knowledge encoded by $\mathbf{e}$ is sufficient to conclude that $\alpha$ holds. To this end, for each propositional formula, we need a scoring function $\phi_{\alpha}:\mathbb{R}^n\rightarrow \mathbb{R}$ such that $\psi_{\alpha}(\mathbf{e})$ indicates whether $\Lambda(\mathbf{e})\models \alpha$. We now study such scoring functions. 

\subsection{Checking the satisfaction of propositional formulas}
Let us consider scoring functions $\psi_{\alpha}:\mathbb{R}^n\rightarrow \mathbb{R}$, for arbitrary propositional formulas $\alpha$, which satisfy the following condition:
\begin{align}
(\psi_{\alpha}(\mathbf{e})>0) \quad\Leftrightarrow\quad (\Lambda(\mathbf{e})\models \alpha)
\end{align}
where we assume that $\Lambda$ is defined as in \eqref{defLambdalogical}. In other words, $\psi_{\alpha}(\mathbf{e})>0$ holds if $\alpha$ is (known to be) true in the epistemic state encoded by $\mathbf{e}$. Let us write $\mathcal{M}(\alpha)$ for the models of a formula $\alpha$, where $\mathcal{M}(\alpha)\subseteq \mathcal{W}$. We find from the definition of $\Lambda$ that $\Lambda(\mathbf{e})\models \alpha$ is equivalent with
$$
\forall \omega \in \mathcal{M}(\neg \alpha)\,.\, \gamma_{p_{\omega}}(\mathbf{e})>0
$$
In other words, to check the satisfaction of a propositional formula $\alpha$, we need a scoring function that allows us to check whether $\gamma_p(\mathbf{e})>0$ for every $p$ in some subset of properties $\mathcal{Q}\subseteq \mathcal{P}$. In particular let us define a scoring function $\gamma_{\mathcal{Q}}$ for every $\mathcal{Q}\subseteq \mathcal{P}$ such that
\begin{align}\label{eqPrincipleGammaQ}
(\gamma_{\mathcal{Q}}(\mathbf{e}) >0) \quad\equiv\quad (\forall p\in\mathcal{Q}. \gamma_{p}(\mathbf{e}) >0)
\end{align}
If this equivalence is satisfied, we will say that the scoring functions are \emph{faithful}.
Then we have:
\begin{align*}
(\psi_{\alpha}(\mathbf{e})>0) \quad\Leftrightarrow\quad (\gamma_{\mathcal{M}(\neg \alpha)}(\mathbf{e}) >0)
\end{align*}
The scoring functions of the form $\psi_{\alpha}$ are thus a special case of scoring functions of the form $\gamma_{\mathcal{Q}}$. For generality, we will study the latter type of scoring functions in the remainder of this section. This also has the advantage that we can stay closer to the setting from Section \ref{secPoolingProperties}. Analogously to $\mathsf{Pos}_p$ and $\mathsf{Neg}_p$, we now define the following regions:
\begin{align*}
\mathsf{Pos}_{\mathcal{Q}} &= \{\mathbf{e} \in X \mid \gamma_{\mathcal{Q}}(\mathbf{e})>0\} \\
\mathsf{Neg}_{\mathcal{Q}} &= \{\mathbf{e} \in X \mid \gamma_{\mathcal{Q}}(\mathbf{e})\leq 0\} 
\end{align*}
Similarly, under the weak semantics, we can consider scoring functions of the form $\gamma'_{\mathcal{Q}}$, with $\mathcal{Q}\subseteq \mathcal{P}$, which are faithfully linked to the scoring functions $\gamma_p$ as follows:
\begin{align}\label{eqLinkGammaPropSetWeak}
(\gamma'_{\mathcal{Q}}(\mathbf{e}) \geq 0) \quad\equiv\quad (\forall p\in\mathcal{Q}. \gamma_{p}(\mathbf{e}) \geq 0)
\end{align}
The corresponding positive and negative regions are defined as follows
\begin{align*}
\mathsf{Pos}'_{\mathcal{Q}} &= \{\mathbf{e} \in X \mid \gamma'_{\mathcal{Q}}(\mathbf{e})\geq 0\} \\
\mathsf{Neg}'_{\mathcal{Q}} &= \{\mathbf{e} \in X \mid \gamma'_{\mathcal{Q}}(\mathbf{e})< 0\} 
\end{align*}
Clearly, if the scoring functions $\gamma_p$ are continuous, then continuous scoring functions of the form $\gamma_{\mathcal{Q}}$ must also exist, as we can simply define $\gamma_{\mathcal{Q}}(\mathbf{e})=\min_{q\in \mathcal{Q}} \gamma_q(\mathbf{e})$, and similar for the weak semantics. Our main focus in the remainder of this section is on the following question: is it possible for the scoring functions $\gamma_{\mathcal{Q}}$ and $\gamma'_{\mathcal{Q}}$ to be linear? In other words, can we use linear scoring functions to check the satisfaction of a propositional formula in the epistemic state encoded by a vector $\mathbf{e}$? This question is important because of the prevalence of linear scoring functions in the classification layer of neural networks. Our results are summarised in Table \ref{tabConditionsReasoning}. One important finding is that linear scoring functions of the form $\gamma_{\mathcal{Q}}$ are not compatible with the strict semantics, regardless of how the embeddings were obtained (and thus also regardless of which pooling operator is considered). For the weak semantics, we find that linear scoring functions are possible with $\diamond_{\mathsf{max}}$ and $\diamond_{\mathsf{had}}$, but crucially, this is only the case if the set of embeddings $X$ is bounded in a suitable way.

\begin{table}
\centering
\begin{tabular}{llcc}
\toprule
\textbf{Pooling Operator} & & \textbf{Semantics} & \textbf{Linear $\gamma_{\mathcal{Q}}$ possible?}\\
\midrule
\multirow{2}{*}{Average} && Strict & \xmark \\
        && Weak & \xmark \\
\midrule
\multirow{2}{*}{Summation} && Strict & \xmark \\
        && Weak & \xmark  \\
\midrule
\multirow{4}{*}{Max-pooling} &\multirow{2}{*}{$X=\mathbb{R}^n$} & Strict & \xmark \\
 && Weak & \xmark \\
 \cmidrule(l){2-4}
&\multirow{2}{*}{$X=]-\infty,z]^n$} & Strict & \xmark \\
 && Weak & \cmark \\
\midrule
\multirow{4}{*}{Hadamard} & \multirow{2}{*}{$X=\mathbb{R}^n$} & Strict & \xmark\\
        && Weak & \xmark \\
        \cmidrule(l){2-4}
&\multirow{2}{*}{$X=[0,+\infty[^n$} & Strict & \xmark\\
        && Weak & \cmark \\
\bottomrule
\end{tabular}
\caption{Summary of results about the possibility of checking entailment of propositional formulas using linear scoring functions. \label{tabConditionsReasoning}}
\end{table}
\subsection{Linear scoring functions under the strict semantics}\label{secLinearStrict}
In the following, we assume that embeddings belong to some region $X\subseteq\mathbb{R}^n$, which we assume to be convex. We make no assumptions about how the embeddings are obtained, requiring only that exhaustiveness \eqref{eqAvgAssumptionEveryCombination} is satisfied, i.e.\ for every epistemic state, there exists some embedding $\mathbf{e}\in X$ which encodes it. We show that scoring functions of the form $\gamma_{\mathcal{Q}}$ can then not be linear. Before showing the main result, we first prove three lemmas.

\begin{lemma}\label{lemmaGeneralLinearImpossible1}
Suppose exhaustiveness \eqref{eqAvgAssumptionEveryCombination} is satisfied.
Let $\mathcal{Q}=\{p_1,...,p_k\}$ be a subset of $\mathcal{P}$. Assume that $\gamma_{\mathcal{Q}}$ is linear and suppose $\mathbf{e}\in X$ is such that $\gamma_{\mathcal{Q}}(\mathbf{e})=0$. For every $\varepsilon>0$ there exists some  $\mathbf{f}\in X$ such that $d(\mathbf{e},\mathbf{f})<\varepsilon$ while $\gamma_{\mathcal{Q}}(\mathbf{f})>0$.
\end{lemma}
\begin{proof}
Note that the linearity of $\gamma_{\mathcal{Q}}$ means that there exists some hyperplane $H$ which separates the regions $\mathsf{Pos}_{\mathcal{Q}}$ and $\mathsf{Neg}_{\mathcal{Q}}$, where $\gamma_{\mathcal{Q}}(\mathbf{e})=0$ means that $\mathbf{e}\in H$. If $\mathbf{e}\in \textit{int}(X)$ then the claim is trivially satisfied, as there exist vectors $\mathbf{f}$ on either side of the hyperplane which are arbitrarily close to $\mathbf{e}$. Now assume $\mathbf{e}\in \delta(X)$. Suppose there were an $\varepsilon>0$ such that for every $\mathbf{f}\in X$ satisfying $d(\mathbf{e},\mathbf{f})<\varepsilon$, it holds that $\gamma_{\mathcal{Q}}(\mathbf{f})\leq 0$. It then follows, given the convexity of $X$, that $H$ is a bounding hyperplane of $X$, and in particular that $X\subseteq \mathsf{Neg}_{\mathcal{Q}}$. This is a contradiction, given that we assumed that \eqref{eqAvgAssumptionEveryCombination} is satisfied.
\end{proof}

\begin{lemma}\label{lemmaGeneralLinearImpossible3}
Suppose exhaustiveness \eqref{eqAvgAssumptionEveryCombination} is satisfied.
Let $\mathcal{Q}=\{p_1,...,p_k\}$ be a subset of $\mathcal{P}$. Assume that $\gamma_{\mathcal{Q}}$ is linear and that $\gamma_{p_1},...,\gamma_{p_k}$ are all continuous. For every $\mathbf{e}\in X$ we have:
$$
(\gamma_{\mathcal{Q}}(\mathbf{e})=0) \quad\Rightarrow\quad (\forall i\in \{1,...,k\} \mid \gamma_{p_i}(\mathbf{e})\geq 0)
$$
\end{lemma}
\begin{proof}
Suppose $\gamma_{\mathcal{Q}}(\mathbf{e})=0$ and suppose $\gamma_{p_i}(\mathbf{e})< 0$ for some $i\in \{1,...,k\}$. Because $\gamma_{p_i}$ is continuous, there must be some $\varepsilon>0$ such that $\gamma_{p_i}(\mathbf{f})< 0$ for every $\mathbf{f}$ satisfying $d(\mathbf{e},\mathbf{f})<\varepsilon$. However, by Lemma \ref{lemmaGeneralLinearImpossible1} we know that there must be such an $\mathbf{f}$ for which $\gamma_{\mathcal{Q}}(\mathbf{f})> 0$, which implies $\gamma_{p_i}(\mathbf{f})> 0$, a contradiction.
\end{proof}

\begin{lemma}\label{lemmaGeneralLinearImpossible2}
Suppose exhaustiveness \eqref{eqAvgAssumptionEveryCombination} is satisfied.
Let $\mathcal{Q}=\{p_1,...,p_k\}$ be a subset of $\mathcal{P}$. Assume that $\gamma_{\mathcal{Q}}$ is linear and that $\gamma_{p_1},...,\gamma_{p_k}$ are all continuous.  For every $\mathbf{e}\in X$ we have:
$$
(\gamma_{\mathcal{Q}}(\mathbf{e})=0)  \Rightarrow (\exists i\in \{1,...,k\}\,.\, \gamma_{p_i}(\mathbf{e})=0)
$$
\end{lemma}
\begin{proof}
Suppose $\gamma_{\mathcal{Q}}(\mathbf{e})=0$. Then there must be some $i\in \{1,...,k\}$ such that $\gamma_{p_i}(\mathbf{e})\leq 0$, by definition of $\gamma_{\mathcal{Q}}$. From Lemma \ref{lemmaGeneralLinearImpossible3} we furthermore find $\gamma_{p_i}(\mathbf{e})\geq 0$. It follows that $\gamma_{p_i}(\mathbf{e})= 0$.
\end{proof}

\begin{figure}
\centering
\includegraphics[width=150pt]{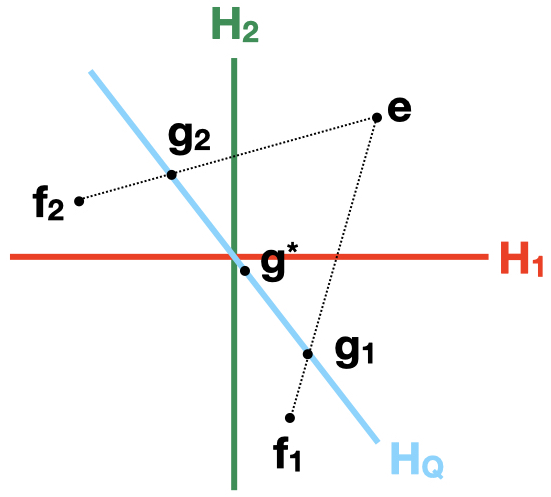}
\caption{Illustration of the construction from the proof of Proposition \ref{propStrictThenNonLinear}. \label{figProofStrictThenNonLinear}}
\end{figure}

\begin{proposition}\label{propStrictThenNonLinear}
Suppose exhaustiveness \eqref{eqAvgAssumptionEveryCombination} is satisfied.
Let $\mathcal{Q}=\{p_1,...,p_k\}$ be a subset of $\mathcal{P}$. Suppose $\gamma_{p_1},...,\gamma_{p_k}$ and $\gamma_{\mathcal{Q}}$ are all linear. Then we have $\vert \mathcal{Q} \vert=1$.
\end{proposition}
\begin{proof}
Figure \ref{figProofStrictThenNonLinear} illustrates the argument provided in this proof.
Suppose $k>1$.
Let $H_{\mathcal{Q}}$ be the hyperplane defined by $H_{\mathcal{Q}} = \{\mathbf{e} \mid \gamma_{\mathcal{Q}}(\mathbf{e})=0\}$. Let $H_i$ similarly be the hyperplane corresponding to $\gamma_{p_i}$. 
Because of \eqref{eqAvgAssumptionEveryCombination} there exists some $\mathbf{e}\in X$ such that $\Gamma(\mathbf{e})=\{p_1,...,p_k\}$. Moreover, for each $i\in \{1,...,k\}$ there exists some $\mathbf{f}_i\in X$ such that $\Gamma(\mathbf{f_i})=\{p_1,...,p_{i-1},p_{i+1},...,p_k\}$. Then we have $\gamma_{\mathcal{Q}}(\mathbf{e})>0$ and $\gamma_{\mathcal{Q}}(\mathbf{f_i})\leq 0$ for every $i\in\{1,...,k\}$. For each $i$, we let $\mathbf{g_i}$ be the point on the intersection between $H_{\mathcal{Q}}$ and the line defined by $\mathbf{e}$ and $\mathbf{f_i}$. Note that $\mathbf{g_i}$ must exist, since $\gamma_{\mathcal{Q}}(\mathbf{e})>0$ and $\gamma_{\mathcal{Q}}(\mathbf{f_i})\leq 0$. Moreover, we have that  $\mathbf{g_i}\in X$ since $X$ is convex. Since $\gamma_{p_1},...,\gamma_{p_k}$ are linear, the fact that $\gamma_{p_j}(\mathbf{e})>0$ and $\gamma_{p_j}(\mathbf{f_i})>0$, for $j\neq i$, implies that $\gamma_{p_j}(\mathbf{g_i})>0$. Since this holds for every $j\neq i$ while $\gamma_{\mathcal{Q}}(\mathbf{g_i})=0$ it follows from Lemma \ref{lemmaGeneralLinearImpossible2} that $\gamma_{p_i}(\mathbf{g_i})=0$. Now consider $\mathbf{g^*}=\frac{1}{k}(\mathbf{g_1}+...+\mathbf{g_k})$. By the convexity of $X$, we have $\mathbf{g^*}\in X$. Moreover, since $\gamma_{p_i}(\mathbf{g_j})>0$ if $i\neq j$ and $\gamma_{p_i}(\mathbf{g_j})=0$ otherwise, if $k>1$ we have that $\gamma_{p_i}(\mathbf{g^*})>0$ for every $i\in\{1,...,k\}$. This would mean that $\Gamma(\mathbf{g^*})=\{p_1,...,p_k\}$ and thus that we should have $\gamma_{\mathcal{Q}}(\mathbf{g^*})>0$. However, since $\mathbf{g_1},...,\mathbf{g_k}\in H_{\mathcal{Q}}$, we also have $\mathbf{g^*}\in H_{\mathcal{Q}}$ and thus $\gamma_{\mathcal{Q}}(\mathbf{g^*})=0$, a contradiction. It follows that $k=1$.
\end{proof}

\subsection{Linear scoring functions under the weak semantics}
In Section \ref{secLinearStrict}, we found that the strict semantics is not compatible with the use of linear scoring functions of the form $\gamma_{\mathcal{Q}}$. We now explore whether linear scoring functions can be more successful under the weak semantics. We already know from Section \ref{secPoolingProperties} that $\diamond_{\mathsf{avg}}$ and $\diamond_{\mathsf{sum}}$ are not compatible with the use of continuous scoring functions under the weak semantics. We therefore focus on the remaining pooling operators, although we will return to $\diamond_{\mathsf{avg}}$ in Section \ref{secReasoningWithAverages}.

\subsubsection{Max-pooling}
For the max-pooling operator $\diamond_{\mathsf{max}}$, it is possible to satisfy faithfulness \eqref{eqLinkGammaPropSetWeak} using linear scoring functions when choosing $X=]-\infty,z]^n$. For instance, let us fix $X=]-\infty,0]^n$. Let $\mathcal{P}=\{p_1,...,p_n\}$. For $p_i\in\mathcal{P}$ we define:
\begin{align*}
\gamma_{p_i}(e_1,...,e_n) = e_i
\end{align*}
Note that, because we assumed $X=]-\infty,0]^n$, it holds that $\gamma_{p_i}(e_1,...,e_n)\geq 0$ iff $e_i=0$. As in Section \ref{secPoolingPropertiesMax}, with this choice we find that the epistemic pooling principle \eqref{eqEpistemicPoolingWeak} is satisfied for all $\mathbf{e},\mathbf{f}\in\mathbb{R}^n$ while exhaustiveness \eqref{eqAvgAssumptionEveryCombinationWeak} is also satisfied. For $\mathcal{Q}=\{p_{i_1},...,p_{i_k}\}$ we now define:
\begin{align*}
\gamma_{\mathcal{Q}}(e_1,...,e_n) = e_{i_1} + ... + e_{i_k}
\end{align*}
Then we have $\gamma_{\mathcal{Q}}(e_1,...,e_n)\geq 0$ iff $e_{i_1} = ... = e_{i_k}=0$, which is equivalent to requiring $\gamma_{p_{i_1}}(e_1,...,e_n)\geq 0,...,\gamma_{p_{i_k}}(e_1,...,e_n)\geq 0$. We thus find that faithfulness \eqref{eqLinkGammaPropSetWeak} is indeed satisfied. 

We now show that faithfulness \eqref{eqLinkGammaPropSetWeak} cannot be satisfied with linear scoring functions if we choose $X=\mathbb{R}^n$. In Proposition \ref{propMaxPoolingBoxes} we found that $\textit{cl}(\mathsf{Neg}_p)$ is of the form $Y_p^1\times ...\times Y_p^n$ (for $X=\mathbb{R}^n$), where for every $i\in\{1,...n\}$ we have $Y_p^i=]-\infty,b_i]$ for some $b_i\in \mathbb{R}$ or $Y_p^i=]-\infty,+\infty[$. While this result was shown for the strict semantics, the same argument can be made for the weak semantics, i.e.\ if the epistemic pooling principle \eqref{eqEpistemicPoolingWeak} is satisfied for every $\mathbf{e},\mathbf{f}\in \mathbb{R}^n$, then we have that $\textit{cl}(\mathsf{Neg}'_p)$ is of the form $Y_p^1\times ...\times Y_p^n$. Clearly, a region $\mathsf{Neg}'_p$ of this form can only arise from a linear scoring function $\gamma_p$ if $Y_p^i=]-\infty,+\infty[$ for all but one $i$ from $\{1,...n\}$. Note that there must exist some $i\in\{1,...,n\}$ such that $Y_p^i$ is of the form $]-\infty,b_i]$ to avoid the trivial case where $\mathsf{Neg}'_p=\mathbb{R}^n$, which would imply that epistemic states in which $p$ is known cannot be modelled.  If $\gamma_p$ is linear, for $p\in \mathcal{P}$, we thus find that there must exist some $i\in \{1,...,n\}$ and some $b_i\in \mathbb{R}$ such that
\begin{align}\label{eqReasoningWeakMax1}
\mathsf{Neg}'_p = \{(e_1,...,e_n) \mid e_i < b_i\}
\end{align}
Note that we have a strict inequality in \eqref{eqReasoningWeakMax1} since we know that $\mathsf{Neg}'_p$ is an open set if $\gamma_p$ is continuous. Given this characterisation of the regions $\mathsf{Neg}'_p$, we now easily find that scoring functions of the form $\mathsf{Neg}'_{\mathcal{Q}}$ can only be linear in the trivial case where $\vert \mathcal{Q} \vert = 1$.

\begin{proposition}\label{propMaxPoolingDisjunctionLinear}
Suppose that the epistemic pooling principle \eqref{eqEpistemicPoolingWeak} is satisfied for every $\mathbf{e},\mathbf{f}\in\mathbb{R}^n$, with $\diamond=\diamond_{\mathsf{max}}$. Suppose exhaustiveness \eqref{eqAvgAssumptionEveryCombinationWeak} is satisfied.
Let $\mathcal{Q}=\{p_1,...,p_k\}$ be a subset of $\mathcal{P}$. Suppose $\gamma_{p_1},...,\gamma_{p_k}$ and $\gamma_{\mathcal{Q}}$ are all linear. Then we have $\vert \mathcal{Q} \vert=1$.
\end{proposition}
\begin{proof}
We know that for each property $p_i$, there exists an index $\ell_i\in\{1,...,n\}$ and a corresponding threshold $b_i\in \mathbb{R}$ such that:
$$
\mathsf{Neg}'_{p_i} = \{(e_1,...,e_n) \mid e_{\ell_i} < b_i\}
$$
and thus we also have:
$$
\mathsf{Pos}'_{p_i} = \{(e_1,...,e_n) \mid e_{\ell_i} \geq b_i\}
$$
Note that $\mathsf{Pos}'_{\mathcal{Q}} = \mathsf{Pos}'_{p_1} \cap ... \cap \mathsf{Pos}'_{p_n}$. This means:
$$
\mathsf{Pos}'_{\mathcal{Q}} = \{(e_1,...,e_n) \mid e_{\ell_1} \geq b_1,...,e_{\ell_k} \geq b_k\}
$$
Clearly, $\mathsf{Pos}'_{\mathcal{Q}}$ can only be linearly separated from $\mathsf{Neg}'_{\mathcal{Q}}$ if $\ell_1=...=\ell_k$. Given \eqref{eqAvgAssumptionEveryCombinationWeak}, this is only possible if $p_1=...=p_k$.
\end{proof}

Going beyond linear scoring functions, it is possible to satisfy faithfulness \eqref{eqLinkGammaPropSetWeak} even with $X=\mathbb{R}^n$. To see this, let $\mathcal{P}=\{p_1,...,p_n\}$ and let us define:
$$
\gamma_{p_i}(e_1,...,e_n) = -\text{ReLU}(-e_i)
$$
with $\text{ReLU}(a)=\max(0,a)$ the rectified linear unit. Note that $\gamma_{p_i}(e_1,...,e_n)\geq 0$ iff $\text{ReLU}(-e_i)=0$ iff $e_i\geq 0$. With this choice, we thus have that $\mathsf{Neg}'_{p_i}$ is of the form \eqref{eqReasoningWeakMax1}. It is easy to verify that \eqref{eqEpistemicPoolingWeak} is satisfied for every $\mathbf{e},\mathbf{f}\in\mathbb{R}^n$, and that \eqref{eqAvgAssumptionEveryCombinationWeak} is also satisfied.
For $\mathcal{Q}=\{p_{i_1},...,p_{i_k}\}$ we define:
\begin{align*}
\gamma_{\mathcal{Q}}(e_1,...,e_n) = -\text{ReLU}(-e_{i_1}) - ... -\text{ReLU}(-e_{i_k})
\end{align*}
Then we have $\gamma_{\mathcal{Q}}(e_1,...,e_n)\geq 0$ iff $\mathsf{ReLU}(-e_{i_1})=...=\mathsf{ReLU}(-e_{i_k})=0$, which is the case if and only if $e_{i_1}\geq 0,...,e_{i_k}\geq 0$, which is equivalent with $\gamma_{p_{i_1}}(e_1,...,e_n)\geq 0,...,\gamma_{p_{i_k}}(e_1,...,e_n)\geq 0$. We thus find that faithfulness \eqref{eqLinkGammaPropSetWeak} is indeed satisfied. Note that through the use of the $\text{ReLU}$ activation function, we essentially end up in the setting where $X=]-\infty,0]^n$, where linear scoring functions are possible.

\subsubsection{Hadamard product}
For the Hadamard product $\diamond_{\mathsf{had}}$ we find that linear scoring functions can be used under the weak semantics, provided that we choose $X=[0,+\infty[^n$. To see this, let $\mathcal{P}=\{p_1,...,p_n\}$ and let the scoring functions $\gamma_{p_i}$ be defined as follows:
\begin{align*}
\gamma_{p_i}(e_1,...,e_n) = -e_i
\end{align*}
It is straightforward to verify that the epistemic pooling principle \eqref{eqEpistemicPoolingWeak} is then satisfied for every $\mathbf{e},\mathbf{f}\in [0,+\infty[^n$, while exhaustiveness \eqref{eqAvgAssumptionEveryCombinationWeak} is also satisfied.
For $\mathcal{Q}=\{p_{i_1},...,p_{i_k}\}$ we define:
\begin{align*}
\gamma_{\mathcal{Q}}(e_1,...,e_n) = -e_{i_1}-...-e_{i_k}
\end{align*}
Then, for $(e_1,...,e_n)\in [0,+\infty[^n$, we have $\gamma_{\mathcal{Q}}(e_1,...,e_n)\geq 0$ iff $e_{i_1}=...=e_{i_k}=0$ iff $\gamma_{p_{i_\ell}}\geq 0$ for every $\ell\in\{1,...,k\}$. Thus we find that faithfulness \eqref{eqLinkGammaPropSetWeak} is indeed satisfied.

In Section \ref{secPoolingPropertiesHad}, we also considered the case where $X=\mathbb{R}^n$. Unfortunately, for this choice of $X$, scoring functions of the form $\gamma_{\mathcal{Q}}$ can only be linear in the trivial case where $\vert \mathcal{Q} \vert = 1$. 
\begin{proposition}\label{propHadPoolingDisjunctionLinear}
Suppose that the epistemic pooling principle \eqref{eqEpistemicPoolingWeak} is satisfied for every $\mathbf{e},\mathbf{f}\in\mathbb{R}^n$, with $\diamond=\diamond_{\mathsf{had}}$. Suppose exhaustiveness \eqref{eqAvgAssumptionEveryCombinationWeak} is satisfied.
Let $\mathcal{Q}=\{p_{i_1},...,p_{i_k}\}$ be a subset of $\mathcal{P}$. Suppose $\gamma_{p_{i_1}},...,\gamma_{p_{i_k}}$ and $\gamma_{\mathcal{Q}}$ are all linear. Then we have $\vert \mathcal{Q} \vert=1$.
\end{proposition}
\begin{proof}
Let $H_{\mathcal{Q}}$ be the hyperplane associated with $\gamma_{\mathcal{Q}}$, i.e.\ $H_{\mathcal{Q}}=\{\mathbf{e}\mid \gamma_{\mathcal{Q}}(\mathbf{e})=0\}$, and let $H_1,...,H_k$ similarly be the hyperplanes associated with $\gamma_{p_{i_1}},...,\gamma_{p_{i_k}}$. Clearly, for each $\mathbf{e}\in H_{\mathcal{Q}}$ we have $\gamma_{p_{i_1}}(\mathbf{e})\geq 0,...,\gamma_{p_{i_k}}(\mathbf{e})\geq 0$. This is only possible if the hyperplanes $H_{\mathcal{Q}},H_1,...,H_k$ are all parallel. Given that we assumed that \eqref{eqAvgAssumptionEveryCombinationWeak} is satisfied, this is only possible if $p_{i_1}=...=p_{i_k}$.
\end{proof}
Similar as we found for max-pooling, it is possible to satisfy faithfulness \eqref{eqLinkGammaPropSetWeak} for $X=\mathbb{R}^n$ by using a non-linear activation function. In this case, the aim of this activation function is to map all vectors $\mathbf{e}\in \mathbb{R}^n$ to some vector in $[0,+\infty[^n$, where we know that linear scoring functions are possible. In particular, we can use the following:
\begin{align*}
\gamma_{p_i}(e_1,...,e_n) &= -e_i^2\\
\gamma_{\mathcal{Q}}(e_1,...,e_n) &= -e_{i_1}^2-...-e_{i_k}^2
\end{align*}
It is easy to verify that faithfulness \eqref{eqLinkGammaPropSetWeak} is indeed satisfied for this choice, while the epistemic pooling principle \eqref{eqEpistemicPoolingWeak} is satisfied for every $\mathbf{e},\mathbf{f}\in\mathbb{R}^n$ and exhaustiveness \eqref{eqAvgAssumptionEveryCombinationWeak} is also satisfied.

\subsection{Reasoning with average pooling}\label{secReasoningWithAverages}
The results we have established so far suggest that neural networks are inherently limited in their reasoning abilities when averaging (or summation) is used. Graph Neural Networks often rely on this pooling operator, even when they are used in applications where they are expected to learn to carry out reasoning tasks. In this section, we analyse whether the empirical success of GNNs in such settings can be explained by weakening the epistemic pooling principle.

In Section \ref{secLinearStrict} we found that linear scoring functions cannot be used under the strict semantics. Note that this limitation holds regardless of how the embeddings were obtained. Moreover, the limitation arises as soon as we have two properties $p_1$ and $p_2$, such that we need to be able to check whether $p_1$ holds, whether $p_2$ holds, and whether $p_1$ and $p_2$ both hold. In Section \ref{secRealizationAverage} we found that continuous scoring functions cannot be used under the weak semantics for $\diamond_{\mathsf{avg}}$. Since the limitation holds for arbitrary continuous scoring functions, adding a (continuous) non-linearity after pooling cannot resolve the issue. Hence, there does not appear to exist a straightforward mechanism to use average pooling while also satisfying the epistemic pooling principle, which seems to be at odds with the prevalence and empirical success of models that rely on average pooling. One possible argument is that the epistemic pooling principle is perhaps too restrictive. While it should not be the case that $\mathbf{e} \diamond_{\mathsf{avg}} \mathbf{f}$ captures properties that are not captured by $\mathbf{e}$ or $\mathbf{f}$, we only need this principle to hold for the embeddings $\mathbf{e},\mathbf{f}$ that we are likely to encounter in practice. In particular, due to the way neural networks are trained, embeddings satisfying some property $p$ are typically separated \emph{by some margin} from embeddings which do not. Formally, let $\Delta>0$ represent a given margin. Then we can expect that embeddings $\mathbf{e}$ will either satisfy $\gamma_p(\mathbf{e})\leq 0$ or  $\gamma_p(\mathbf{e})\geq \Delta$. If this is the case for every property $p$, we can think of $\mathbf{e}$ as representing a clear-cut epistemic state. In contrast, when $0 < \gamma_p(\mathbf{e}) < \Delta$, we can think of $\mathbf{e}$ as being ambiguous regarding the satisfaction of $p$. Let us write $X^*\subseteq X$ for the clear-cut epistemic states, i.e.
\begin{align}
X^* = \{\mathbf{e} \in X \mid \forall p\in \mathcal{P}\,.\, \gamma_p(\mathbf{e})\leq 0 \vee \gamma_p(\mathbf{e})\geq \Delta\}
\end{align}
We show that it is possible to satisfy the epistemic pooling principle for all embeddings from $X$, while using scoring functions that are linear after the application of a \text{ReLU} activation, provided that we only require the scoring functions to satisfy faithfulness \eqref{eqPrincipleGammaQ} for embeddings $\mathbf{e}$ in $X^*$. 

Let $X=[0,+\infty[^n$, $\Delta>0$ and $\mathcal{P}=\{p_1,...,p_n\}$. Let us define the scoring functions $\gamma_p$ as follows. 
\begin{align*}
\gamma_{p_i}(e_1,...,e_n)= e_i
\end{align*}
As in Section \ref{secRealizationAverage}, we find that with this choice, the epistemic pooling principle \eqref{eqEpistemicPoolingGeneral} is satisfied for all $\mathbf{e},\mathbf{f}\in X$ while exhaustiveness  \eqref{eqAvgAssumptionEveryCombination} is also satisfied. Note furthermore that $X^*= (\{0\}\cup [\Delta,+\infty[)^n$. For $\mathcal{Q}=\{p_{i_1},...,p_{i_k}\}$ we define:
\begin{align*}
\gamma_{\mathcal{Q}}(e_1,...,e_n) = \Delta - \sum_{\ell=1}^k \text{ReLU}(\Delta-e_{i_{\ell}})
\end{align*}
Then \eqref{eqPrincipleGammaQ} is satisfied for any $(e_1,...,e_n)\in X^*$. Indeed, for $(e_1,...,e_n)\in X^*$ we find:
\begin{align*}
\gamma_{\mathcal{Q}}(e_1,...,e_n) > 0 
&\Leftrightarrow \sum_{\ell=1}^k \text{ReLU}(\Delta-e_{i_{\ell}}) < \Delta\\
&\Leftrightarrow \forall \ell\in \{1,...,k\}\,.\, e_{i_{\ell}}\geq \Delta\\
&\Leftrightarrow \forall \ell\in \{1,...,k\}\,.\, \gamma_{p_{i_{\ell}}}(e_1,...,e_n)>0
\end{align*}
where we used the fact that either $\text{ReLU}(\Delta-e_{i_{\ell}})=0$ or $\text{ReLU}(\Delta-e_{i_{\ell}})=\Delta$, given that $(e_1,...,e_n)\in X^*$. Note that the \text{ReLU} transformation converts $(e_1,...,e_n)$ into a vector that is binary, in the sense that each coordinate is either 0 or $\Delta$. Using the sigmoid function $\sigma$ we can similarly ensure that faithfulness \eqref{eqPrincipleGammaQ} is satisfied for any $(e_1,...,e_n)\in X^*$, by constructing vectors that are approximately binary in the aforementioned sense. For instance, we could define:
\begin{align*}
\gamma_{\mathcal{Q}}(e_1,...,e_n) = \mu - \sum_{\ell=1}^k \sigma\left(\lambda\left(\frac{\Delta}{2}-e_{i_{\ell}}\right)\right)
\end{align*}
where $\mu$ is an arbitrary constant satisfying $0<\mu<1$ and $\lambda>0$ is chosen in function of $\mu$. In particular, by choosing a sufficiently large value for $\lambda$, we can always ensure that
$$
\sigma\left(\lambda\cdot \frac{\Delta}{2}\right) \geq \mu
$$
and 
$$
\sigma\left(-\lambda\cdot \frac{\Delta}{2}\right)< \frac{\mu}{k}
$$
We have $\gamma_{\mathcal{Q}}(e_1,...,e_n)>0$ iff
\begin{align*}
\sum_{\ell=1}^k \sigma\left(\lambda\left(\frac{\Delta}{2}-e_{i_\ell}\right)\right)  < \mu
\end{align*}
Given the assumption that $\lambda$ is sufficiently large, and given that $(e_1,...,e_n)\in X^*$, this inequality is satisfied iff $e_{i_\ell}\geq \Delta$ for every $\ell\in\{1,...,k\}$. The rest of the argument then follows as before.

We can also use linear scoring functions, by defining $X^*$ such that embeddings are approximately binary. In particular, let $X=[0,1]^n$ with $\Delta=1-\varepsilon$ for some $\varepsilon\in ]0,\frac{1}{k}[$. Let $\gamma_{p_j}(e_1,...,e_n)=e_j$ as before. For $\mathcal{Q}=\{p_{i_1},...,p_{i_k}\}$, we define:
$$
\gamma_{\mathcal{Q}}(e_1,...,e_n) = \left(\sum_{\ell=1}^k e_{i_{\ell}}\right) - k + 1 
$$
We then have $\gamma_{\mathcal{Q}}(e_1,...,e_n)>0$ iff
\begin{align*}
\sum_{\ell=1}^k e_{i_{\ell}} > k-1
\end{align*}
Given our assumption that $\varepsilon< \frac{1}{k}$, this inequality is satisfied iff $e_{i_{\ell}}\geq \Delta = 1-\varepsilon$ for every $\ell\in\{1,...,k\}$. The rest of the argument then follows as before.

Note that in each case, we either restrict the setting to embeddings that are approximately binary, or we apply a non-linear transformation to convert the initial embeddings into approximately binary ones. In this way, we replaced the need for discontinuous scoring functions by the assumption that embeddings essentially encode discrete states. It remains unclear whether models that are trained using gradient descent based optimisation can learn to reason under such conditions.
\section{Modelling weighted epistemic states}\label{secWeightedEpistemicStates}

We now consider a setting in which each property $p$ from $\mathcal{P}$ is associated with a certainty level from $\Lambda=\{0,1,...,K\}$. Intuitively, a certainty level of 0 means that we know nothing about $p$, whereas a level of $K$ means that we are fully certain that $p$ is true. When aggregating evidence from different sources, we assume that the certainty level of $p$ is given by the certainty level of the most confident source, in accordance with possibility theory \cite{DBLP:books/sp/DuboisP88}. This setting has the advantage that we can continue to view pooling operators in terms of accumulating knowledge.

\paragraph{Reduction to the standard setting}
It is possible to model weighted epistemic states within the standard framework we have considered so far, using scoring functions of the form $\gamma_{\mathcal{Q}}$. We consider the strict semantics here, but an entirely similar argument can be made for the weak semantics. Let us define the set of extended properties as follows:
$$
\mathcal{P}^\sharp = \{p\sharp i \mid p\in\mathcal{P}, i\in \Lambda\}
$$
For each $p\sharp i\in \mathcal{P}^\sharp$, we let $\gamma_{p\sharp i}$ be a scoring function satisfying \eqref{eqEpistemicPoolingGeneral} for all embeddings $\mathbf{e},\mathbf{f}\in X$. We can interpret $p\sharp i$ as encoding that the certainty level of $p$ is not equal to $i$. For $p\in \mathcal{P}$ and $i\in \Lambda \setminus \{0\}$ we define:
\begin{align*}
\mathcal{Q}_{(p,i)} = \{ p\sharp j \mid p\in \mathcal{P}, j\in \Lambda, j<i\}
\end{align*}
Let $\gamma_{\mathcal{Q}_{(p,i)}}$ be a scoring function satisfying \eqref{eqPrincipleGammaQ}. Then $\gamma_{\mathcal{Q}_{(p,i)}}(\mathbf{e})>0$ means that all certainty levels below $i$ can be excluded for $p$, i.e.\ that $p$ is certain at least to degree $i$.

\paragraph{Weighted epistemic pooling principle}
The construction from the previous paragraph has an important drawback: the number of properties is increased $(K+1)$-fold. From Section \ref{secPoolingProperties}, we know that this implies that the number of dimensions also has to increase $(K+1)$-fold. If for every $p$ and $i$, we need the ability to model that the certainty of $p$ is not $i$, this increase in dimensionality is inevitable. However, in practice, we are typically not interested in excluding arbitrary sets of certainty degrees, only in establishing lower bounds on certainty degrees. To study this setting, we introduce a generalisation of the epistemic pooling principle to weighted epistemic states. Let us write $\langle p,i\rangle$ for the fact that the certainty level of $p$ is at least $i$, where $\langle p,0 \rangle$ means that nothing is known about $p$ whereas $\langle p,K \rangle$ means that $p$ is known with full certainty. We write $\Lambda_0=\{1,2,...,K\}$ for the set of non-trivial lower bounds. We furthermore assume that the certainty level of $p$ is determined by the scoring function $\gamma_p:\mathbb{R}^n\rightarrow \mathbb{R}$. In particular, under the strict semantics, we assume that $p$ is known with certainty at least $i$, for $i\in \Lambda_0$, if $\gamma_{p}>i-1$. 
We then consider the following generalisation of the epistemic pooling principle:
\begin{align}\label{eqAvgPoolingDegrees}
\forall p\in\mathcal{P}. \forall i\in \Lambda_0. (\gamma_{p}(\mathbf{e})>i-1) \vee (\gamma_{p}(\mathbf{f})>i-1) \Leftrightarrow \gamma_{p}(\mathbf{e}  \diamond \mathbf{f}) >i-1
\end{align}
We define $\Gamma_{\Lambda}$ as a set of weighted properties, encoding for each property what is the highest certainty degree with which this property is believed:
\begin{align}
\Gamma_{\Lambda}(\mathbf{e}) = \{\langle p, \max(0,\min(K,\lceil \gamma_p(\mathbf{e})\rceil))\rangle \mid p\in \mathcal{P}\}
\end{align}
Exhaustiveness, i.e.\ the condition that every weighted epistemic state is modelled by some vector $\mathbf{e}\in X$ can then be formalised as follows:
\begin{align}\label{eqEveryWeightedEpistemicState}
\forall \mu:\mathcal{P} \rightarrow \Lambda \,.\, \exists \mathbf{e}\in X\,.\, \Gamma_{\Lambda}(\mathbf{e}) = \{\langle p,\mu(p) \rangle \mid p\in \mathcal{P}\}
\end{align}
In other words, for every assignment of certainty degrees to the properties in $\mathcal{P}$, there exists an embedding $\mathbf{e}\in X$ that encodes the corresponding epistemic state. We also introduce the following notations for $i\in \Lambda_0$:
\begin{align*}
\mathsf{Pos}_{p,i} &= \{\mathbf{e}\in X\mid \gamma_p(\mathbf{e})>i-1\}\\
\mathsf{Neg}_{p,i} &= \{\mathbf{e}\in X\mid \gamma_p(\mathbf{e})\leq i-1\} = X \setminus \mathsf{Pos}_{p,i}
\end{align*}

Similarly, we can also consider a weighted version of the weak epistemic pooling principle:
\begin{align}\label{eqAvgPoolingDegreesWeak}
\forall p\in\mathcal{P}. \forall i\in \Lambda_0. (\gamma_{p}(\mathbf{e})\geq i-1) \vee (\gamma_{p}(\mathbf{f})\geq i-1) \Leftrightarrow \gamma_{p}(\mathbf{e}  \diamond \mathbf{f}) \geq i-1
\end{align}
The weighted epistemic state associated with an embedding $\mathbf{e}\in X$ is now defined as follows:
\begin{align}
\Gamma'_{\Lambda}(\mathbf{e}) = \{\langle p, \max(0,\min(K,1 + \lfloor \gamma_p(\mathbf{e})\rfloor))\rangle \mid p\in \mathcal{P}\}
\end{align}
while the counterpart of exhaustiveness \eqref{eqEveryWeightedEpistemicState} becomes
\begin{align}\label{eqEveryWeightedEpistemicStateWeak}
\forall \mu:\mathcal{P} \rightarrow \Lambda \,.\, \exists \mathbf{e}\in X\,.\, \Gamma'_{\Lambda}(\mathbf{e}) = \{\langle p,\mu(p) \rangle \mid p\in \mathcal{P}\}
\end{align}
We now analyse whether the weighted epistemic pooling principles \eqref{eqAvgPoolingDegrees} and \eqref{eqAvgPoolingDegreesWeak} can be satisfied in a non-trivial way for embeddings with $\vert\mathcal{P}\vert$ dimensions. We find that this is only the case for $\diamond_{\mathsf{max}}$, as shown in Table \ref{tabConditionsEpistemicWeighted}.

\begin{table}
\centering
\begin{tabular}{lcc}
\toprule
\textbf{Pooling Operator}  & \textbf{Semantics} & \textbf{$n=|\mathcal{P}|$ possible?} \\
\midrule
\multirow{2}{*}{Average} & Strict & \xmark \\
        & Weak & \xmark \\
\midrule
\multirow{2}{*}{Summation} & Strict & \xmark \\
        & Weak & \xmark  \\
\midrule
\multirow{2}{*}{Max-pooling} & Strict & \cmark \\
 & Weak & \cmark \\
\midrule
\multirow{2}{*}{Hadamard}  & Strict & \xmark\\
        & Weak & \xmark\\
\bottomrule
\end{tabular}
\caption{Summary of results about the realizability of the weighted epistemic pooling principles for embeddings of dimensionality $n=|\mathcal{P}|$. \label{tabConditionsEpistemicWeighted}}
\end{table}

\subsection{Realizability of the weighted epistemic pooling principle}

\paragraph{Average}

In entirely the same way as in Section \ref{secRealizationAverage}, we then find that $\mathsf{Pos}_{p,i}$ and $\mathsf{Neg}_{p,i}$ are convex, for every $p\in\mathcal{P}$ and $i\in \Lambda_0$. Similarly, we also find that $\mathsf{Neg}_{p,i}\subseteq \delta(X)$ for every $p\in\mathcal{P}$ and $i\in \Lambda_0$. Since $\mathsf{Pos}_{p,i}\setminus \mathsf{Pos}_{p,K}\subseteq \mathsf{Neg}_{p,K}$, we thus also have that $\mathsf{Pos}_{p,i}\setminus \mathsf{Pos}_{p,K}\subseteq \delta(X)$. We can now show the following result using a similar strategy as in the proof of Proposition \ref{propDimAvg1}. 

\begin{proposition}\label{propDimAvg1WeightedWeighted}
Suppose the weighted epistemic pooling principle \eqref{eqAvgPoolingDegrees} is satisfied for all embeddings $\mathbf{e},\mathbf{f}\in X$, with $\diamond=\diamond_{\mathsf{avg}}$ and $X\subseteq \mathbb{R}^n$. Suppose that exhaustiveness \eqref{eqEveryWeightedEpistemicState} is satisfied. It holds that $n\geq \vert\mathcal{P}\vert \cdot K$.
\end{proposition}
\begin{proof}
Let $p_1,...,p_{\vert\mathcal{P}\vert}$ be an enumeration of the properties in $\mathcal{P}$. 
Note that because of \eqref{eqEveryWeightedEpistemicState}, we have that $\mathsf{Neg}_{p_1,K}\neq \emptyset$ and $\mathsf{Pos}_{p_1,K}\neq\emptyset$. Since both of these regions are convex, it follows from the hyperplane separation theorem that there exists a hyperplane $H_{1,K}$ which separates $\mathsf{Neg}_{p_1,K}$ and $\mathsf{Pos}_{p_1,K}$. Since $\mathsf{Neg}_{p_1,K}\subseteq \delta(X)=\textit{cl}(\mathsf{Pos}_{p_1,K})$, we furthermore know that $\mathsf{Neg}_{p_1,K}\subseteq H_{1,K}$.

Note that $H_{1,K}\cap \mathsf{Neg}_{p_1,K-1}$ and $H_{1,K}\cap \mathsf{Pos}_{p_1,K-1}$ are convex regions. Moreover, since $\mathsf{Neg}_{p_1,K}\subseteq H_{1,K}$, $\mathsf{Neg}_{p_1,K}\cap \mathsf{Neg}_{p_1,K-1}\neq \emptyset$, and  $\mathsf{Neg}_{p_1,K}\cap \mathsf{Pos}_{p_1,K-1}\neq \emptyset$, we find that $H_{1,K}\cap \mathsf{Neg}_{p_1,K-1}\neq \emptyset$ and $H_{1,K}\cap \mathsf{Pos}_{p_1,K-1}\neq\emptyset$. It follows from the hyperplane separation theorem that there exists some hyperplane $H_{1,K-1}$ separating $H_{1,K}\cap \mathsf{Neg}_{p_1,K-1}$ and $H_{1,K}\cap \mathsf{Pos}_{p_1,K-1}$. Moreover, it holds that $H_{1,K}\cap \mathsf{Neg}_{p_1,K-1} \subseteq H_{1,K-1}$ (which follows in the same way as $H_1\cap \mathsf{Neg}_{p_2} \subseteq H_2$ in the proof of Proposition \ref{propDimAvg1}). In particular, we then also have that $\mathsf{Neg}_{p_1,K} \cap \mathsf{Neg}_{p_1,K-1} \subseteq H_{1,K-1}$. We can repeat this argument to show that for each $i\in \{0,...,K-1\}$ there is a hyperplane $H_{1,K-i}$ separating $H_{1,K}\cap...\cap H_{1,K-i+1}\cap \mathsf{Neg}_{p_1,K-i}$ and $H_{1,K}\cap...\cap H_{1,K-i+1}\cap \mathsf{Pos}_{p_1,K-i}$, such that $H_{1,K}\cap...\cap H_{1,K-i+1}\cap \mathsf{Neg}_{p_1,K-i}\subseteq H_{1,K-i}$. This implies that $H_{1,K}\cap ... \cap H_{1,K-i+1}\not\subseteq H_{i,K-i}$ and thus $\textit{dim}(H_{1,K}\cap...\cap H_{1,K-i}) = n-i-1$. In particular, we have $\textit{dim}(H_{1,K}\cap...\cap H_{1,1}) = n-K$.

Due to \eqref{eqEveryWeightedEpistemicState}, we have that $H_{1,K}\cap...\cap H_{1,1}\cap \mathsf{Neg}_{p_2,K}\neq \emptyset$ and $H_{1,K}\cap...\cap H_{1,1} \cap \mathsf{Pos}_{p_2,K}\neq\emptyset$. It follows that there exists a hyperplane $H_{2,K}$ separating $H_{1,K}\cap...\cap H_{1,1}\cap \mathsf{Neg}_{p_2,K}$ and $H_{1,K}\cap...\cap H_{1,1} \cap \mathsf{Pos}_{p_2,K}$, such that $H_{1,K}\cap...\cap H_{1,1}\cap \mathsf{Neg}_{p_2,K}\subseteq H_{2,K}$, from which it follows that $\textit{dim}(H_{1,K}\cap...\cap H_{1,1} \cap H_{2,K}) = n-K-1$. Continuing as before, we find $\textit{dim}(H_{1,K}\cap...\cap H_{1,1} \cap H_{2,K}...\cap H_{2,1}) =n-2K$. Repeating this argument for every property in $\mathcal{P}$, we find:
$$
\textit{dim}\left(\bigcap_{j=1}^{\vert\mathcal{P}\vert} \bigcap_{i=1}^k H_{j,i}\right) =n-\vert\mathcal{P}\vert \cdot K.
$$
Since this is only possible if $n-\vert\mathcal{P}\vert \cdot K\geq 0$ it follows that $n\geq \vert\mathcal{P}\vert \cdot K$.
\end{proof}
In Section \ref{secRealizationAverage}, we found that the epistemic pooling principle cannot be satisfied in a non-trivial way under the weak semantics, if continuous scoring functions are used. This result carries over to the weighted setting.

\paragraph{Summation}
Entirely similar as in Section \ref{secRealisabilitySummation}, we find that when the weighted epistemic pooling principle is satisfied for all $\mathbf{e},\mathbf{f}\in X$ and $\diamond=\diamond_{\mathsf{sum}}$, then it is also satisfied for $\diamond=\diamond_{\mathsf{avg}}$. It thus follows from Proposition \ref{propDimAvg1WeightedWeighted} that $\vert\mathcal{P}\vert \cdot K$ dimensions are needed to satisfy the weighted epistemic pooling principle \eqref{eqAvgPoolingDegrees} in a non-trivial way. Moreover, we also have that the weighted epistemic pooling principle under the weak semantics, i.e.\ the weak weighted epistemic pooling principle \eqref{eqAvgPoolingDegreesWeak}, cannot be satisfied when continuous scoring functions are used.

\paragraph{Max-pooling}
For $\diamond_{\mathsf{max}}$, we can satisfy the weighted epistemic pooling principles \eqref{eqAvgPoolingDegrees} and \eqref{eqAvgPoolingDegreesWeak} for every $\mathbf{e},\mathbf{f}\in \mathbb{R}^n$, while ensuring that every weighted epistemic state is encoded by some vector. Let $\mathcal{P}=\{p_1,...,p_n\}$ and let $\gamma_{p_i}$ be defined as follows:
\begin{align*}
\gamma_{p_i}(e_1,...,e_n)= e_i
\end{align*}
It holds that the strict weighted epistemic pooling principle \eqref{eqAvgPoolingDegrees} is satisfied for each $\mathbf{e},\mathbf{f}\in \mathbb{R}^n$. Indeed, for every $j\in \Lambda_0$ we have:
\begin{align*}
&(\gamma_{p_i}(e_1,...,e_n)>j-1) \vee (\gamma_{p_i}(f_1,...,f_n)>j-1)\\
&\quad\quad\Leftrightarrow (e_i > j-1) \vee (f_i > j-1)\\
&\quad\quad\Leftrightarrow \max(e_i,f_i) > j-1\\
&\quad\quad\Leftrightarrow \gamma_{p_i}((e_1,...,e_n)\diamond_{\mathsf{max}} (f_1,...,f_n))>j-1
\end{align*}
In entirely the same way, we find that the weak weighted epistemic pooling principle \eqref{eqAvgPoolingDegreesWeak} is satisfied.
We now show that exhaustiveness is satisfied, i.e.\ \eqref{eqEveryWeightedEpistemicState} and  \eqref{eqEveryWeightedEpistemicStateWeak}. Let $\mu:\mathcal{P}\rightarrow \Lambda$. Then we can define $\mathbf{e}=(e_1,...,e_n)$ as follows:
\begin{align*}
e_i = \mu(p_i) - \frac{1}{2}
\end{align*}
It is trivial to verify that $\Gamma_{\Lambda}(\mathbf{e})= \Gamma'_{\Lambda}(\mathbf{e})= \{\langle p,\mu(p) \rangle \mid p\in \mathcal{P}\}$.

\paragraph{Hadamard product}
As in Section \ref{secPoolingPropertiesHad}, we find that $\mathsf{Pos}_{p,\ell}$ is a finite union of regions of the form $X \cap \bigcap_{i\in I} H_i$, with $H_i=\{(x_1,...,x_n)\in\mathbb{R}^n \mid x_i=0\}$. For a given index set $I\subseteq \{1,...,n\}$, we define:
$$
\mathcal{P}^{\Lambda}_I = \{(p,\ell)\in \mathcal{P}\times \Lambda_0 \mid X \cap \bigcap_{i\in I} H_i \subseteq \mathsf{Pos}_{p,\ell}\}
$$
In entirely the same way as in Lemma \ref{lemmaHadIndexSet2}, we find for every $I,J\subseteq \{1,...,n\}$ that:
\begin{align}\label{eqLemmaHadIndexSet2Weighted}
\mathcal{P}^{\Lambda}_{I\cup J} = \mathcal{P}^{\Lambda}_{I} \cup \mathcal{P}^{\Lambda}_{J}
\end{align}
The following lemma is also shown in entirely the same way as Lemma \ref{lemmaHadIndexSet1}.
\begin{lemma}\label{lemmaHadIndexSet1Weighted}
Let $X=\mathbb{R}^n$ or $X=[0,+\infty[^n$. Suppose the weighted epistemic pooling principle \eqref{eqAvgPoolingDegrees} is satisfied for all embeddings $\mathbf{e},\mathbf{f}\in X$, with $\diamond=\diamond_{\mathsf{had}}$. Let $p\in \mathcal{P}$ and $\ell\in\Lambda_0$ such that $\mathbf{e}=(e_1,...,e_n) \in\mathsf{Pos}_{p,\ell}$. Let $I=\{i\in \{1,...,n\} \mid e_i=0\}$. It holds that
$$
X \cap \bigcap_{i\in I} H_i \subseteq \mathsf{Pos}_{p,\ell}
$$
\end{lemma}

We can then generalise Proposition \ref{propMinDimHadStrict} as follows.
\begin{proposition}\label{propMinDimHadStrictWeighted}
Let $X=\mathbb{R}^n$ or $X=[0,+\infty[^n$. 
Suppose the weighted epistemic pooling principle \eqref{eqAvgPoolingDegrees} is satisfied for all embeddings $\mathbf{e},\mathbf{f}\in X$, with $\diamond=\diamond_{\mathsf{had}}$. Suppose that exhaustiveness \eqref{eqEveryWeightedEpistemicState} is satisfied. It holds that $n\geq \vert\mathcal{P}\vert \cdot K$.
\end{proposition}
\begin{proof}
Given \eqref{eqEveryWeightedEpistemicState}, for each $p\in \mathcal{P}$ and $\ell\in \Lambda_0$, there must exist some $(e^{p,\ell}_1,...,e^{p,\ell}_n)\in X$ such that $\Gamma(e^{p,\ell}_1,...,e^{p,\ell}_n)=\{\langle p,\ell\rangle\} \cup \{\langle q,0\rangle \mid q\in \mathcal{P}\setminus \{p\}\}$. Let us fix such vectors $(e^{p,\ell}_1,...,e^{p,\ell}_n)$ for each $p\in \mathcal{P}$ and $\ell\in \Lambda_0$. Define $I_{p,\ell} = \{i\in\{1,...,n\} \mid e^{p,\ell}_i=0\}$. Note that by Lemma \ref{lemmaHadIndexSet1Weighted}, we have $X\cap \bigcap_{i\in I_{p,\ell}}H_i \subseteq \mathsf{Pos}_{p,\ell}$. Moreover, by construction, we have $X\cap \bigcap_{i\in I_{p,\ell}}H_i \not\subseteq \mathsf{Pos}_{q,\ell'}$ for any $q\neq p$ and $\ell'\in\Lambda_0$, and similarly, we have $X\cap \bigcap_{i\in I_{p,\ell}}H_i \not\subseteq \mathsf{Pos}_{p,\ell'}$ for any $\ell'>\ell$. In other words, we have that all elements in $\mathcal{P}^{\Lambda}_{I_{p,\ell}}$ are of the form $(p,\ell')$ with $\ell'\leq \ell$. 
For $p\neq q$ and $\ell,\ell'\in \Lambda_0$, we clearly have $I_{p,\ell}\not\subseteq I_{q,\ell'}$, since $I_{p,\ell}\subseteq I_{q,\ell'}$ would imply $\langle p,\ell\rangle \in \mathcal{P}_{I_{q,\ell'}}$. 
This means in particular that $I_{p,\ell}\neq \emptyset$ for every $p\in \mathcal{P}$ and $\ell\in\Lambda_0$. 

Let $(q_1,\ell_1),...,(q_s,\ell_s)$, with $s=\vert\mathcal{P}\vert \cdot K$, be an enumeration of the elements from $\mathcal{P}\times \Lambda_0$, such that whenever $\ell<\ell'$, it holds that $(p,\ell)$ is listed before $(p,\ell')$, for any $p\in\mathcal{P}$.
Note that we have already established that $I_{q_1,\ell_1}\neq \emptyset$.
Now we show that for any given $k\in\{2,...,s\}$, it holds that $I_{q_k,\ell_k}\not\subseteq I_{q_1,\ell_1}\cup I_{q_2,\ell_2}\cup ... \cup I_{q_{k-1},\ell_{k-1}}$. Indeed $I_{q_k,\ell_k}\subseteq I_{q_1,\ell_1}\cup I_{q_2,\ell_2}\cup ... \cup I_{q_{k-1},\ell_{k-1}}$ would imply $(q_k,\ell_k)\in \mathcal{P}^{\Lambda}_{I_{q_1,\ell_1}\cup ...\cup I_{q_{k-1},\ell_{k-1}}}$. From \eqref{eqLemmaHadIndexSet2Weighted}, we know that $\mathcal{P}^{\Lambda}_{I_{q_1,\ell_1}\cup ...\cup I_{q_{k-1},\ell_{k-1}}}= \mathcal{P}^{\Lambda}_{I_{q_1,\ell_1}}\cup ... \cup \mathcal{P}^{\Lambda}_{I_{q_{k-1},\ell_{k-1}}}$ and we know that the latter can only contain elements of the form $(q_1,\ell_1'), ...,(q_{k-1},\ell_{k-1}')$ with $\ell_1'\leq \ell_1,...,\ell_{k-1}'\leq \ell_{k-1}$. Due to the assumption we made that  $(p,\ell)$ is listed before $(p,\ell')$ whenever $\ell<\ell'$, we thus find  $(q_k,\ell_k)\notin \mathcal{P}^{\Lambda}_{I_{q_1,\ell_1}\cup ...\cup I_{q_{k-1},\ell_{k-1}}}$, a contradiction. Hence, we have $I_{q_k,\ell_k}\not\subseteq I_{q_1,\ell_1}\cup I_{q_2,\ell_2}\cup ... \cup I_{q_{k-1},\ell_{k-1}}$. This means there is at least one element in $I_{(q_k,\ell_k)}$ which does not occur in $I_{q_1,\ell_1},...,I_{q_{k-1},\ell_{k-1}}$. Since this needs to hold for every $k\in\{2,...,s\}$, there need to be at least $s$ distinct elements in $I_{q_1,\ell_1}\cup ...\cup I_{q_1,\ell_s}$. In particular, we thus have that $n\geq s = \vert\mathcal{P}\vert \cdot K$.
\end{proof}
The above limitation also applies to the weak semantics. However, if we choose $X=[0,1]^n$ it is possible to satisfy the weighted epistemic pooling principle \eqref{eqAvgPoolingDegrees} for $n$ properties if $K=2$ (i.e.\ if we have three certainty degrees). Indeed, let $\mathcal{P}=\{p_1,...,p_n\}$. Then we can define:
\begin{align}\label{eqWeightedHad01Construction}
\gamma_{p_i}(e_1,...,e_n) = 
\begin{cases}
\frac{3}{2} & \text{if $e_i=0$}\\
-\frac{1}{2} & \text{if $e_i=1$}\\
\frac{1}{2} & \text{otherwise}
\end{cases}
\end{align}
Then it is straightforward to verify that the weighted epistemic pooling principle \eqref{eqAvgPoolingDegrees} is indeed satisfied for every $\mathbf{e},\mathbf{f}\in [0,1]^n$ and that \eqref{eqEveryWeightedEpistemicState} also holds. The above construction also provides an example of how the weighted epistemic pooling principle can be satisfied for the weak semantics. Indeed, with the above definition of $\gamma_{p_i}$ we have that the weak weighted epistemic pooling principle \eqref{eqAvgPoolingDegreesWeak} is satisfied for every $\mathbf{e},\mathbf{f}\in [0,1]^n$, while exhaustiveness \eqref{eqEveryWeightedEpistemicStateWeak} also holds.
In Section \ref{secPoolingPropertiesHad}, we found that the weak epistemic pooling principle could be satisfied with continuous scoring functions when $\diamond=\diamond_{\mathsf{had}}$. Unfortunately, this strategy does not allow us to obtain a continuous alternative to the scoring functions defined in \eqref{eqWeightedHad01Construction}.

\section{Discussion and related work}\label{secRelatedWork}

\subsection{Logical reasoning with neural networks}
The use of neural networks for simulating symbolic reasoning has been extensively studied under the umbrella of neuro-symbolic reasoning \cite{DBLP:series/faia/BesoldGBBDHKLLPPPZ21}. The seminal KBANN method \cite{towell1994knowledge}, for instance, uses feedforward networks with carefully chosen weights to simulate the process of reasoning with a given rule base. In this case, the neural network simulates a fixed rule base, which is manually specified. More recent work has investigated how a neural network can be trained to simulate the deductive closure of a given logical theory, e.g.\ a description logic ontology \cite{DBLP:journals/jair/HoheneckerL20}. The idea that rule-based reasoning can be simulated using neural networks also lies at the basis of various strategies for learning rules from data \cite{DBLP:conf/nips/Rocktaschel017,DBLP:conf/nips/YangYC17,DBLP:conf/nips/SadeghianADW19}. Another recent research line has focused on whether standard neural network architectures, such as LSTMs or transformer based language models \cite{BERT}, can be trained to recognise logical entailment \cite{DBLP:conf/iclr/EvansSAKG18,DBLP:conf/ijcai/ClarkTR20}. In these works, the input consists of a premise (or a set of premises) and a hypothesis, and the aim is to predict whether the hypothesis can be inferred from the premise(s). The aforementioned works differ from this paper, as our focus is not on whether neural networks can simulate logical reasoning, but on whether vectors can be used for encoding epistemic states. Simulating logical reasoning does not necessarily require that vectors can encode epistemic states, since we can treat reasoning as an abstract symbol manipulation problem. Moreover, the scope of what we address in this paper goes beyond logical reasoning, as the epistemic pooling principle also matters whenever we need to combine evidence from different sources (e.g.\ features being detected in different regions of an image).

This paper builds on our earlier work \cite{DBLP:conf/akbc/Schockaert21}, where the focus was on the following question: given a set of attributes $\mathcal{A}$, a pooling function $\diamond$ and a propositional knowledge base $K$, can we always find an embedding $\mathbf{a}$ and a scoring function $\gamma_a$ for every $a\in \mathcal{A}$ such that: 
$$
\gamma_b(\mathbf{a_1}\diamond ... \diamond \mathbf{a_k})\geq 0 \quad \Leftrightarrow \quad K\cup \{a_1,...,a_k\} \models b
$$
for any $a_1,...,a_k,b\in \mathcal{A}$. This analysis fundamentally differs from the results in this paper, because in the case of \cite{DBLP:conf/akbc/Schockaert21}, we only care about the behaviour of the pooling operator and scoring function for a finite set of vectors, i.e.\ the attribute embeddings. In the notations of this paper, this amounts to limiting $X$ to a finite set of carefully chosen embeddings. As a result, for instance, in \cite{DBLP:conf/akbc/Schockaert21} it was possible to use average pooling in combination with continuous scoring functions, under the weak semantics, something which we found to be impossible in the more general setting considered in this paper. 

One may wonder whether the approach we take in this paper is too strict, e.g.\ whether it is really necessary to insist that the epistemic pooling principle is satisfied for all embeddings. Note, however, that during training, the embeddings will change after each update step of the (gradient-descent based) optimizer. If we want the model to learn to reason, then it is important that all of these embeddings can be viewed as epistemic states.  In other words, if pooling is only meaningful for a particular discrete set of vectors, then we may end up with an architecture that is capable of reasoning in theory, but where those parameters that would lead to meaningful behaviour cannot be learned in practice.

\subsection{Reasoning with Graph Neural Networks}
In Section \ref{secPropositionalReasoning}, we have specifically focused on reasoning in the context of propositional logic. However, our analysis is relevant for reasoning in relational domains as well. A standard approach for relational reasoning with neural networks is to rely on Graph Neural Networks (GNNs) \cite{scarselli2008graph,wu2020comprehensive}.  Given a graph $G=(V,E)$, with $V$ a set of nodes and $E\subseteq V\times V$ a set of edges, a GNN aims to learn a vector representation of every node in $V$. This is achieved by incrementally updating the current representation of each node based on the representations of their neighbours. In particular, let us write $\mathbf{v^{(i)}}$ for the represention of node $v\in V$ in layer $i$ of the GNN. Let $\{u_1,...,_k\}$ be the set of neighbours of $v$ in $G$. Then the representation of $v$ in layer $i+1$ is typically computed as follows:
\begin{align}\label{eqDefGNN}
\mathbf{v^{(i+1)}} = f_2(\mathbf{v^{(i)}} \diamond_1 f_2(f_3(\mathbf{u_1^{(i)}}) \diamond_2 ... \diamond_2 f_3(\mathbf{u_k^{(i)}})))
\end{align}
The functions $f_1$, $f_2$, $f_3$, $\diamond_1$ and $\diamond_2$ can be defined in various way. However, regardless of the specifics, we can think of $\diamond_1$ and $\diamond_2$ as pooling operators, whereas $f_1$, $f_2$ and $f_3$ correspond to (possibly non-linear) transformations. Note that we used a non-standard notation in \eqref{eqDefGNN} to highlight the connection to this paper. Intuitively, we can think of $f_3(\mathbf{u_j^{(i)}})$ as a vector that captures what we can infer about the entity represented by node $v$ from the fact that it is connected to $u_j$. In multi-relational settings (e.g.\ knowledge graphs), where different types of edges occur, $f_3$ can be replaced by a function that depends on the edge type. The pooling operator $\diamond_2$ is used to aggregate the evidence coming from the different neighbours of $v$, whereas $\diamond_1$ is used to combine the evidence we already have about $v$ with the evidence we can obtain from its neighbours. 

The ability of GNNs to simulate logical reasoning has been studied in \cite{DBLP:conf/iclr/BarceloKM0RS20}. The idea is that each node is associated with a set of properties, which are referred to as colours. We can then consider first-order formulas involving unary predicates, referring to these colours, and the binary predicate $E$, which captures whether two nodes are connected. For instance, consider the following formula:
$$
\phi_1(x) \equiv \textit{Green}(x) \wedge (\exists y\,.\, E(x,y) \wedge \textit{Blue}(y)) \wedge \neg(\exists y\,.\, E(x,y) \wedge \textit{Red}(y))
$$
This formula is true for a given node if it is green and it is connected in the graph to a blue node but not to a red node. We can also consider counting quantifiers, as in the following example:
$$
\phi_2(x) \equiv (\exists^{\geq 5} y\,.\, E(x,y) \wedge \textit{Blue}(y)) 
$$
This formula is true for a given node if it is connected to at least 5 blue nodes. The question studied in  \cite{DBLP:conf/iclr/BarceloKM0RS20} is which formulas can be recognised by a GNN, i.e.\ for which class of formulas $\phi$ can we design a GNN such that we can predict whether $\phi$ holds for a node $n$ from the final-layer embedding of that node, using some scoring function. In particular, it was shown that the set of formulas that can be recognised (without global read-out) are exactly those that are expressible in graded modal logic \cite{DBLP:journals/sLogica/Rijke00}, which is characterised as follows:
\begin{enumerate}
\item for each color $C$, the formula $C(x)$ is a graded modal logic formula;
\item if $\phi(x)$ and $\psi(x)$ are graded modal logic formulas and $n\in \mathbb{N}$, then the following formulas are also graded modal logic formulas: $\neg \phi(x)$, $\phi(x)\wedge \psi(x)$ and $\exists^{\geq n} y\,.\, E(x,y) \wedge \phi(y)$.
\end{enumerate}
The proof that was provided for the characterisation in \cite{DBLP:conf/iclr/BarceloKM0RS20} is constructive. It relies on the particular choice of $\diamond_2$ as summation, which appears to be at odds with the limitations that were identified for $\diamond_{\mathsf{sum}}$ in this paper. However, the GNN in their construction only uses binary coordinates. As we have seen in Section \ref{secReasoningWithAverages}, in that case, $\diamond_{\mathsf{avg}}$ can be used for propositional reasoning, a result which straightforwardly carries over to $\diamond_{\mathsf{sum}}$. This observation may help to explain the discrepancy between the theoretical ability of GNNs to capture arbitrary formulas from graded logic, and the challenges that have empirically been observed when using GNNs for learning to reason. For instance, GNNs have generally failed to outperform simpler embedding based methods for the task of knowledge graph completion \cite{ali2020benchmarking}, while \cite{sinha-etal-2019-clutrr} found that GNNs were limited in their ability to generalise in a systematic way from examples that were more complex than the ones seen during training. 

Interestingly, \cite{cucala2021explainable} recently proposed a GNN for knowledge graph completion in which $\diamond_2$ corresponds to max-pooling. Together with a number of other design choices (e.g.\ avoiding negative weights and using a particular encoding of the knowledge graph), this leads to GNNs that are in some sense equivalent to a set of rules. The suitability of max-pooling, in this context, is not a surprise, given our results from Section \ref{secPropositionalReasoning}. Note, however, that our results also suggest that coordinates have to be upper-bounded if we want to identify cases where sets of atomic properties are jointly satisfied. In the approach from \cite{cucala2021explainable}, this issue is avoided by using high-dimensional embeddings in which each candidate inference corresponds to a separate coordinate, which amounts to treating the formulas of interest as atomic properties in our framework. 

\subsection{Modelling relations as regions}
A popular strategy for making predictions in relational domains consists in learning (i) an embedding $\mathbf{e}\in\mathbb{R}^n$ for each entity of interest $e$ and (ii) scoring function $f_r: \mathbb{R}^n\times\mathbb{R}^n \rightarrow \mathbb{R}$ for each relation $r$ such that $f_r(\mathbf{e},\mathbf{f})$ indicates the probability that the fact $r(e,f)$ holds. Despite the fact that such methods intuitively carry out some form of logical inference, and despite their strong empirical performance \cite{ali2020benchmarking}, for most approaches, there is no clear link between the parameters of the model (i.e.\ the embeddings and the parameters of the scoring functions), on the one hand, and the kinds of inferences that are captured, on the other hand. Region-based methods, however, are a notable exception \cite{DBLP:conf/kr/Gutierrez-Basulto18,DBLP:conf/nips/AbboudCLS20,DBLP:conf/iclr/RenHL20,zhang2021cone,DBLP:journals/corr/abs-2206-04192}.
The central idea of such methods is to represent predicates as regions. For instance, if $s$ is a unary predicate, then the corresponding region $R_s$ is such that $\mathbf{e}\in R_s$ iff $s(e)$ is true. For a binary predicate $r$, one option is to use a region $R_r$ in $\mathbb{R}^{2n}$ such that $\mathbf{e}\oplus\mathbf{f} \in R_r$ iff $r(e,f)$ is true. In other words, we view the concatenation of $\mathbf{e}$ and $\mathbf{f}$ as the embedding of the tuple $(e,f)$ and model binary predicates as regions over such concatenations. The key advantage of region-based models is that logical relationships can be directly encoded, in terms of spatial relationships between the region-based representations of the predicates involved. As a simple example, the rule $r_1(x,y)\rightarrow r_2(x,y)$ is satisfied if $R_{r_1} \subseteq R_{r_2}$. We refer to \cite{DBLP:conf/kr/Gutierrez-Basulto18} for details on how more complex rules can be similarly captured. This correspondence between logical dependencies and spatial relationships can be used to ensure that the predictions of the model are in accordance with a given knowledge base, or to explain the behaviour of a model in terms of the logical rules it captures. Moreover, region based embeddings make it possible to query embeddings of knowledge bases in a principled way \cite{DBLP:conf/iclr/RenHL20,zhang2021cone}. Essentially, a given query (e.g.\ ``retrieve all companies whose headquarter is in a European capital city'') is then mapped onto a region, such that the entities that satisfy the query are those whose embedding belongs to the region.

However, it should be noted that the aforementioned region-based embeddings encode a specific possible world, rather than an epistemic state. In other words, they encode which facts are assumed to be true and false, but they cannot encode incomplete knowledge (e.g.\ that either $r(a,b)$ or $s(a,b)$ holds). 
In \cite{DBLP:journals/amai/LeemhuisOW22} a geometric model based on cones was presented, which has the ability to encode incomplete knowledge to some extent. For instance, these geometric models can capture the fact that it is unknown whether some entity $e$ belongs to some concept $A$. Essentially, each concept is represented by two cones. Entities whose representation belongs to the first cone are those which are known to instances of the concept; entities whose representation belongs to the second cone are those which are known not to be instances of the concept; and the remaining instances are those whose membership is unknown. Note, however, that not all epistemic states can be captured in this way; e.g.\ we cannot represent the fact that either $e$ belongs to $A$ or $f$ belongs to $B$. Along similar lines, the box embeddings proposed in \cite{DBLP:conf/acl/McCallumVLM18,DBLP:conf/iclr/LiVZBM19} can be used to model some epistemic states. For instance, an approach for capturing uncertain knowledge graph embeddings based on box embeddings was proposed in \cite{DBLP:conf/naacl/ChenBCDLM21}. In this case, the idea is to represent the entities themselves as regions, and as axis-aligned hyperboxes in particular. The problem of pooling such hyperbox representations has not yet been considered, to the best of our knowledge. The most intuitive approach would be to simply take the intersection of the boxes, i.e.\ if entity $e$ is represented by a hyperbox $B_1$, according to one source, and by a hyperbox $B_2$, according to another source, we may want to use $B_1\cap B_2$ as an aggregate representation of entity $e$, reflecting the information provided by both sources. However, this leads to a number of practical challenges. For instance, if box embeddings are used to parameterise a probabilistic model, as in \cite{DBLP:conf/naacl/ChenBCDLM21}, then it is unclear whether a sound justification can be provided for a pooling operation that relies intersecting the entity-level box representations. Such probabilistic models also serve a rather different purpose to the framework that we studied in this paper, which is about accumulating knowledge rather than about quantifying uncertainty. Even in settings where the box embeddings are used as a purely qualitative representation, a problem arises when the region $B_1\cap B_2$ is empty. Intuitively, box embeddings act as constraints on possible worlds, and such constraints can be inconsistent. This is different from the settings we studied in this paper, which were about accumulating knowledge, formalised as sets of properties.

\section{Conclusions}
Neural networks are often implicitly assumed to perform some kind of reasoning. In this paper, we have particularly focused on the common situation where evidence is obtained from different sources, which then needs to be combined. The core question we addressed is whether it is possible to represent the evidence obtained from each source as a vector, such that pooling these vectors amounts to combining the corresponding evidence, a requirement we refer to as the \emph{epistemic pooling principle}. This question is important for understanding whether, or under which conditions, neural networks that rely on pooling are able to perform reasoning in a principled way. Our analysis shows that standard pooling operators can indeed be used for accumulating evidence, but only under particular conditions. Broadly speaking, the requirement that the epistemic pooling principle is satisfied substantially limits how knowledge can be encoded. For instance, when average pooling is used, we find that embeddings have to be limited to a strict subset $X$ of $\mathbb{R}^n$, and that vectors which encode that some property $p$ is \emph{not} satisfied have to be located on a bounding hyperplane of $X$. We also highlighted how such conditions limit the way in which embeddings can be used. For instance, when average pooling is used, it is not possible to use linear scoring functions for checking whether a given propositional formula is satisfied in the epistemic state encoded by a given vector. In general, our results provide valuable insights for the design of neural networks that are required to implement some form of systematic reasoning.

\section*{Acknowledgements}
This work was supported by EPSRC grants EP/V025961/1 and EP/W003309/1.

\bibliography{refs}
\bibliographystyle{elsarticle-num}

\end{document}